\documentclass[journal]{IEEEtran}
\usepackage{fixltx2e}
\usepackage[english]{babel}
\usepackage{xcolor}
\usepackage{tablefootnote}
\usepackage{multirow}
\usepackage{amssymb}
\usepackage{float}
\usepackage{longtable}
\usepackage{float}
\usepackage{tabularx}
\usepackage[letterpaper,top=2cm,bottom=2cm,left=3cm,right=3cm,marginparwidth=1.75cm]{geometry}
\usepackage{amsmath}
\usepackage{graphicx}
\usepackage{adjustbox}
\usepackage{wrapfig}
\usepackage[colorlinks=true, allcolors=blue]{hyperref}
\usepackage{caption}
\usepackage{subcaption}
\usepackage{siunitx}
\usepackage{supertabular,booktabs}
\usepackage{afterpage}
\usepackage{makecell}

\begin{document}
%

\title{Deep Imbalanced Multi-Target Regression: 3D Point Cloud Voxel Content Estimation in Simulated Forests}

%
%
%

\author{Amirhossein Hassanzadeh,
        Bartosz Krawczyk,
        Michael Saunders,
        Rob Wible,
        Keith Krause, 
        Dimah Dera, and
        Jan van Aardt
        
        \thanks{This work was supported by the National Geospatial-Intelligence Agency, Award No. HM0476-20-1-000, Project Title: Enhanced 3D Sub-Canopy Mapping via Airborne/Spaceborne Full-Waveform LiDAR. (Corresponding author: Amirhossein Hassanzadeh.)}
        
        \thanks{Amirhossein Hassanzadeh, Michael Saunders, Bartosz Krawczyk, Dimah Dera, and Jan van Aardt are with the Chester F. Carlson Center for Imaging Science, Rochester Institute of Tech- nology, Rochester, NY 14623 USA (e-mail: axhcis@rit.edu)}

        \thanks{Rob Wible is with the United State Space Force (USSF).}
        
        \thanks{Keith Krause is with Battelle.}                
        }

%
%

\markboth{Journal of \LaTeX\ Class Files,~Vol.~13, No.~9, Month~YEAR}%
{Shell \MakeLowercase{\textit{et al.}}: Bare Demo of IEEEtran.cls for Journals}
%



\maketitle


\begin{abstract} 
Voxelization is an effective approach to reduce the computational cost of processing Light Detection and Ranging (LiDAR) data, yet it results in a loss of fine-scale structural information. This study explores whether low-level voxel content information, specifically target occupancy percentage within a voxel, can be inferred from high-level voxelized LiDAR point cloud data collected from  Digital Imaging and remote Sensing Image Generation (DIRSIG) software. In our study, the targets include bark, leaf, soil, and miscellaneous materials. We propose a multi-target regression approach in the context of imbalanced learning using Kernel Point Convolutions (KPConv). Our research leverages cost-sensitive learning to address class imbalance called density-based relevance (DBR). We employ weighted Mean Saquared Erorr (MSE), Focal Regression (FocalR), and regularization to improve the optimization of KPConv. This study performs a sensitivity analysis on the voxel size (0.25 - 2 meters) to evaluate the effect of various grid representations in capturing the nuances of the forest. This sensitivity analysis reveals that larger voxel sizes (e.g., 2 meters) result in lower errors due to reduced variability, while smaller voxel sizes (e.g., 0.25 or 0.5 meter) exhibit higher errors, particularly within the canopy, where variability is greatest. For bark and leaf targets, error values at smaller voxel size datasets (0.25 and 0.5 meter) were significantly higher than those in larger voxel size datasets (2 meters), highlighting the difficulty in accurately estimating within-canopy voxel content at fine resolutions. This suggests that the choice of voxel size is application-dependent. Our work fills the gap in deep imbalance learning models for multi-target regression and simulated datasets for 3D LiDAR point clouds of forests. 



\end{abstract}


%
\IEEEpeerreviewmaketitle

\section{\textbf{Introduction}}

\IEEEPARstart{R}{emote} sensing systems are often used to observe, identify, and assess various phenomena with applications ranging from agriculture, ecology, and forestry to climate change and urban planning management~\cite{weiss2020remote, bhaga2020impacts, lechner2020applications, wellmann2020remote}. Among remote sensing systems, Light Detection and Ranging (LiDAR) and its capability to provide high-fidelity 3D structural information have gained interest in the past two decades~\cite{reutebuch2005light,trujillo2011light,liu2008airborne}. Waveform LiDAR systems, compared to discrete LiDAR systems, resemble a wave with a higher penetrating ability, which provides more detailed information about the target area being studied, especially in terms of sub-canopy penetration and understanding complex vegetation structures~\cite{vaughn2012tree, wallace2020full}. A large and growing body of literature has investigated the use of LiDAR systems for Leaf area index (LAI) estimation~\cite{wang2020estimation,zhang2022evaluation, tang2014deriving,wang2023retrieval}, tree species detection and classification~\cite{qin2022individual, liu2022tree, zhong2022identification,qin2022individual}, canopy modeling~\cite{armston2013direct, hermosilla2013estimation, whitehurst2013characterization, jarron2020detection, fisher2020modelling}, and forest and vegetation structure~\cite{hyde2005mapping, hancock2017measurement, hermosilla2013estimation, anderson2006use, jarron2020detection, larue2020compatibility, filippelli2019comparison, oehmcke2024deep}. However, due to the nature of LiDAR systems, high computational costs, long processing times, and high data storage costs are inevitable~\cite{pirotti2011analysis}.

To address these challenges, voxel-based approaches are effective in reducing computational cost and lowering processing time~\cite{hermosilla2013estimation, li2016generating, nie2017above}. Voxelization not only reduces the amount of pre-processing and storage needed, but also enables the acquisition of higher levels of detail within canopy and understory vegetation, mitigating the impact of off-nadir scan angles~\cite{wang2020influence}. Voxel size could impact the performance of the analysis; an optimized voxel size has a strong positive effect. However, there exists a trade-off between computational efficiency and information loss; The larger the voxel size, the more averaged the representation of the signal within it, leading to higher information loss but lower computational costs.~\cite{wang2020influence, ross2022lidar}. Generating real-world ground truth data for any voxelized grid representation, particularly in large and intricate environments like forests, proves to be an impractical and highly resource-intensive task. Simulation models, on the other hand, have proven to be an effective alternative to address this issue. The RIT's Digital Imaging and Remote Sensing Image Generation (DIRSIG) software \cite{goodenough2012dirsig} is a physics-based radiometric model capable of generating radiometrically accurate discrete and waveform LiDAR data. DIRSIG is backed by ray tracing, and it serves as an invaluable source for creating digital counterparts of real-world remote sensing data. DIRSIG's approach to scene building in terms of accurate geometric identification of objects, such as leaves, trees, and bark, enables precise calculation of each material’s area and, in turn, their voxel contribution. This is particularly powerful, as the generated ground truth data is unattainable in real-world scenarios, and the large volume of generated data enables the use of data-driven models. 

LiDAR systems aim to capture the volumetric structure of 3D environments, and we adopt the regular voxelized grid, or rank 3 tensor, as a versatile and commonly applicable data structure for characterizing such environments. Voxelized representations can be constructed either directly from high-resolution point clouds or by more sophisticated techniques that account for full waveform distributions. 

\smallskip
\noindent \textbf{Research goal.} In this paper we aim to show that it is possible to effectively separate the voxelized LiDAR data of the forest into semantic categories of interest---voxel percentage occupancy of bark, leaf, soil, and miscellaneous---based on voxel spatial and sensor features. 

\smallskip
\noindent \textbf{Main contributions.} This paper offers the following contributions:
\begin{itemize}
\item \textbf{Novel and physics-based dataset.} We construct a new simulated LiDAR dataset using DIRSIG with a versatile discrete LiDAR system over the Harvard Forest site. This dataset is unique in the literature, as its ground truth values are derived from precise geometrical calculations, enabling robust analysis. This dataset accounts for the radiometric component of remote sensing systems, bringing the simulated data as close as possible to its real-world counterpart.

\item \textbf{Dedicated end-to-end deep learning architecture.} We design a deep network architecture based on Kernel Point Convolution (KPConv) that fits the purpose of multi-target objective, by borrowing concepts from point cloud semantic segmentation and adapting them to multi-target regression. We benchmark against state-of-the-art methods, demonstrating the effectiveness of our approach. Based on our literature review, this paper is the first to apply deep learning-based multi-target regression in a forest environment.

\item \textbf{Novel instance weighting technique for imbalanced regression.} We propose a simple yet cost-sensitive weighting technique for handling target imbalance in a multi-target regression problem, called density-based relevance (DBR). DBR is computationally efficient, relying only on histogram calculations, making it useful for large-scale datasets with millions of samples.

\item \textbf{Novel robust loss function for multi-target regression.} We introduce a new loss function tailored to our problem, leveraging cost-sensitive learning for class imbalance using weighted mean square error (MSE) and sample hardness using Focal Regression (FocalR) for harder-to-learn samples. This loss function helps improve performance, especially within tree canopies. 

\item \textbf{Voxel sensitivity analysis.} We evaluate various voxel sizes representing forest structures and their impact on model performance. This analysis provides valuable insights into the trade-offs between resolution, predictive accuracy and computational complexity in forest modeling.
\end{itemize}

To the best of our knowledge, this is the first study to investigate whether voxel content information can be inferred from discrete LiDAR data gathered from a 3D simulated scene, specifically the Harvard Forest site, using deep learning-based multi-target regression models.

\subsection{\textbf{Background}}
%




\smallskip
\noindent\textbf{Digital Imaging and Remote Sensing Image Generation (DIRSIG)}.
With the advent of technology and increased ease of access to computational resources, simulation models have been of significant aid in deriving solutions for challenging problems~\cite{mobley1998hydrolight, goodenough2012dirsig, berk1987modtran, berger2018evaluation}. Simulation models are capable of recreating various settings by exploring the search space, thereby producing a large quantity of data that can be used as proof of concept, calibration, and validation of real-world scenarios. RIT's digital imaging and remote sensing image generation (DIRSIG) software \cite{goodenough2012dirsig} is a physics-based, first-principle radiometric model for generating remotely sensed data that is geometrically and radiometrically accurate ~\cite{brown2005time,schott1999advanced}. DIRSIG is capable of generating passive and active broad-band, spectral, polarized, low-light, and synthetic aperture radar (SAR) images~\cite{brown2005time,burton2002elastic,flusche2010exploiting,gartley2010comparison,ientilucci1998multispectral,schott1992incorporation,schott1999advanced}. DIRSIG takes in scene characteristics, instrument configurations, and atmosphere properties to simulate collected data from sensor platforms. 

Leveraging DIRSIG’s ability to simulate complex scenarios, its outputs can provide critical insights into challenges like voxelization and fractional coverage estimation. The transformation of point cloud data into voxels unavoidably results in information loss. The dilemma to solve here is whether one is able to use the voxelized information at a coarse scale, and infer fine-scale detail (e.g., fractional coverage) with acceptable accuracy. As the term suggests, fractional coverage is the normalized calculated area of a specific material within a bounded region, also known as a voxel. Fractional coverage significantly helps scholars develop a more nuanced understanding of the forest canopy. Moreover, DIRSIG is an invaluable source for quantifying variables, such as fractional coverage, among others, that are otherwise impossible to achieve at a large scale in real-world scenarios. A simple mathematical representation of the introduced concept would be 
\begin{equation} \label{eq:intro:mapping_functions}
F(\text{signal}) =  A_{1}\% + A_{2}\% + A_{3}\% + \cdots,
\end{equation}
where $A_{1}\%$ is the percentage area of material 1 in the voxel. By annotating the $Signal$ as independent variables, and $A_{1}\% \cdots A_{n}\%$ as dependent variables, the objective at hand aligns with a multi-target regression task.

\smallskip
\noindent\textbf{Simulated LiDAR Datasets.}
Real-world 3D point cloud datasets fall within three different categories mainly separated by how the data were collected: Aerial Laser Scanners (ALS), Terrestrial Laser Scanners (TLS), and Mobile Laser Scanners (MLS). ALS is airborne, often captured by drone and aircraft, and is mostly used for forestry and mapping; MLS are either handheld or mounted on moving vehicle with applications in autonomous driving and urban mapping; TLS are stationary and mainly used for indoor mapping and cultural heritage.

Several studies in the literature focused on simulated LiDAR data for modeling forest environments. Notable examples include TreePointCloud~\cite{lin2020novel}, SmartTree~\cite{dobbs2023smart}, LiDARForest~\cite{lu2024lidar}, TreeNet3d~\cite{tang2024treenet3d}, and Boreal3D~\cite{liu2025advancing}, which have been applied to tasks such as forest biomass and carbon mapping. However, while these datasets are geometrically accurate, they do not incorporate radiometric components, functioning purely as 3D representations.

\smallskip
\noindent\textbf{Machine Learning in LiDAR Analysis.}
The availability of large-scale 3D point cloud LiDAR datasets motivates the use of data-driven models. Deep predictive learning models have become the forefront of science in the past decade, effective in understanding complex patterns. The literature abounds with studies taking advantage of artificial intelligence and LiDAR data for semantic segmentation, classification, mapping, and canopy height regression, among others~\cite{shinohara2020fwnet, assmann2021deep, liu2019deep, liao2018deep, lang2022global, shinohara2020semantic, zorzi2019full}. Semantic instance segmentation is the task of predicting a class for each instance of points based on the scene's semantics. In this study, we are specifically interested in semantic segmentation for instance class prediction in the discrete domain (classification), which could be extrapolated to per-instance target value prediction in the continuous domain (regression). Moreover, taking this concept a step further, it can be extended to multi-target regression, based on the logic explained in Equation~\ref{eq:intro:mapping_functions}, with the purpose of per-voxel content estimation.

Semantic segmentation approaches in deep learning can be divided into different categories based on data processing and models' architectures. These categories are: 

\begin{enumerate}
    \item Point-based methods that process the raw unstructured point cloud~\cite{jhaldiyal2023semantic}, with examples such as PointNet, its enhancement PointNet++, and Repsurf~\cite{ran2022surface}. PointNet++ extends PointNet by hierarchically capturing local features at multiple scales to better represent fine-grained geometry~\cite{qi2017pointnet,qi2017pointnet++}.
    \item Voxel-based methods voxelize the point cloud in 3D grids and leverage the use of 3D convolution kernels to identify patterns~\cite{rs12111729}, with applications in autonomous driving and examples such as VoxelNet~\cite{zhou2018voxelnet}.
    \item Projection-based approaches project the 3D space into the the 2D space and proceed with instance segmentation in the lower-dimensional space via 2D convolution networks, taking advantage of projection schemes such as spherical projection~\cite{wu2018squeezeseg}, with examples such as SqueezeSeg and its successors SqueezeSeg v2 and v3~\cite{wu2018squeezeseg, wu2019squeezesegv2, xu2020squeezesegv3}.
    \item Graph-based methods represent the 3D points as nodes in a graph and the parameters of the network that would need to be learned would be the edges~\cite{fei2022comprehensive}. 3D point clouds can be considered non-Euclidean data, as they do not fall within a typical grid space, with proposed models such as DGCNN~\cite{wang2019dynamic} and SPGraph~\cite{landrieu2018large}.
    \item Transformer-based approaches implement an attention mechanism to dynamically focus on the most relevant patterns and relationships in the data, improving feature representation and overall model performance. PointTransformer and its successors, PointTransformer v2 and v3 are transformer-based deep learning models~\cite{wu2022point, zhao2021point, wu2024point}. Transformer-based models tend to require a large amount of data because of their architectures with a large number of parameters. A more comprehensive review of these models can be found in~\cite{tychola2024deep, fei2022comprehensive}.
\end{enumerate}

Kernel Point Convolutions (KPConv) outperformed state-of-the-art models in the segmentation task, taking advantage of spherical kernel point convolutions~\cite{thomaskpconv2019}. KPConv also outperformed other benchmarks for DALES aerial laser scanner (ALS) data with 83\% mean intersection over union~\cite{thomaskpconv2019, varney2020dales}. Recently, the author introduced two updates to the KPConv, namely KPConvD and KPConvX~\cite{thomas2024kpconvx}. KPConvD adopts a lighter depth-wise design, and KPConvX incorporates kernel self-attention mechanisms to prioritize important regions in gridded data. 

Our literature review indicates that deep learning models have been employed for regression tasks using LiDAR point cloud data. Deep learning models were used to estimate forest variables such as biomass~\cite{garcia2015comparison} and to predict vegetation stratum occupancy from airborne LiDAR data~\cite{kalinicheva2022predicting}. Other models have also been proposed for point cloud registration tasks, taking advantage of regression~\cite{gao2023hdrnet}. Based on the literature review, our study is the first of its kind that leverages point cloud deep learning models for a multi-target regression task in a forest environment. 

\smallskip
\noindent\textbf{Data Imbalance.}
Effectively addressing data imbalance is vital to improve the reliability and precision of predictive models. Supervised predictive modeling can be broadly categorized into classification and regression tasks. Data imbalance can occur in both classification and regression scenarios, and, particularly deep learning models often suffer from a lack of gradient updates for rare samples, leading to significant performance drops for those samples~\cite{fernandez2018learning}. Classification and regression could extend from single-target to multi-target tasks.

For single-output imbalance classification, popular approaches include random oversampling (RO), random undersampling (RU), Synthetic Minority Oversampling Technique (SMOTE) and its variants~\cite{chawla2002smote, dablain2022deepsmote}, Adaptive Synthetic Sampling (ADASYN~\cite{he2008adasyn}), and cost-sensitive learning that incorporates a higher cost for misclassifying rare samples. Metrics such as F1-score, Geometric Mean~\cite{barandela2003strategies}, and Matthews Correlation Coefficient (MCC;~\cite{baldi2000assessing}) are among the evaluation measures used for imbalanced classification. 

For single-output regression, methods such as SMOTE-R (SMOTE for regression;~\cite{torgo2013smote}), Synthetic Minority Oversampling by Introduction of Gaussian Noise (SMOGN;~\cite{branco2017smogn}), resampled bagging for imbalance regression (REBAGG;~\cite{branco2018rebagg}) were introduced. As SMOTE-R and SMOGN both rely on multiple distance metric calculations, it is considered computationally expensive for large datasets. Another study introduced density-based weighting that calculates per sample weight using kernel density estimation (KDE)~\cite{yang2021delving}. Another study proposed a solution for deep models with the introduction of feature distribution smoothing (FDS) and label distribution smoothing (LDS) to compensate for missing target values by convolving Gaussian kernels~\cite{yang2021delving}. With FDS, the study highlighted the importance of feature statistics in balanced datasets and showed that, in imbalanced datasets, the priors of target values learned by the model tend to be similar to those of the majority samples, i.e., biased priors. On the metric side, for the regression of single output imbalance, similar challenges arise. Standard regression metrics, such as $R^2$, do not accurately evaluate the performance of the model. Utility-based measures, such as precision, recall, and F-measure, have been introduced for imbalanced regression~\cite{torgo2013smote}. Torgo \textit{et al.} introduced a relevance function $ \phi : Y \rightarrow [0, 1]$ that assigns relevance scores on a scale of 0 to 1 to the domain of the target variable, where 1 signifies the highest relevance, and 0 represents the lowest relevance~\cite{torgo2007utility} that could be used to calculate the weighted version of regression metrics. 

In the domain of imbalance single-target regression, several Python packages have been developed to address these challenges. SMOGN implements a Synthetic Minority Over-sampling Technique (SMOTE) adapted for regression by adding Gaussian noise~\cite{branco2017smogn}. ResReg provides specialized resampling methods designed specifically for regression tasks~\cite{gado2020improving}. PyImbalReg offers a broader collection of resampling and weighting techniques to handle imbalanced regression datasets~\cite{branco2019pre}. Additionally, ImbalancedLearningRegression implements a variety of strategies aimed at improving model performance on imbalanced regression problems~\cite{wu2022imbalancedlearningregression}. While primarily developed for classification tasks, the algorithms in imbalanced-learn can also be adapted for regression settings~\cite{JMLR:v18:16-365}.


In multi-output classification and regression, the problem becomes even more complex. Among approaches for multi-target regression are the problem transformation with distance-based learning~\cite{hamalainen2020problem}, adaptation methods that extend single-target regression to handle multiple targets~\cite{rodriguez2022rotation}, and ensemble-based regressors aiming to adapt single-output methods to handle multiple targets~\cite{reyes2018ensemble}. The literature shows a scarcity of imbalanced multi-target regression studies. As such, developing robust methodologies to handle imbalance in regression tasks is essential to ensure the fidelity and generalizability of models applied to LiDAR-derived data, allowing for more accurate and comprehensive interpretations of the complex environmental structures.

\begin{figure*}
    \centering
    \subfloat[\centering]{{\includegraphics[width=0.45\textwidth]{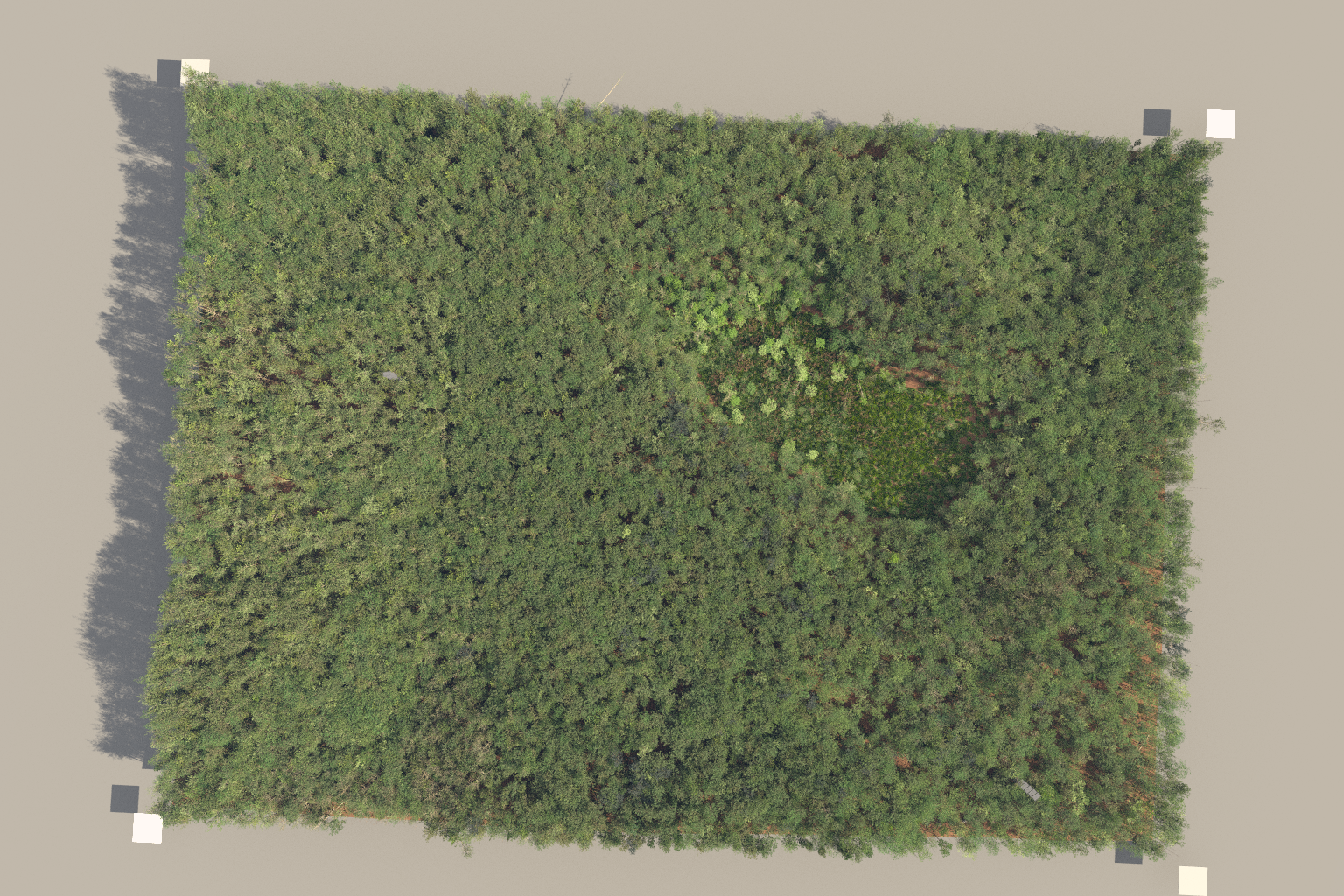} }}
    \subfloat[\centering]{{\includegraphics[width=0.25\textwidth]{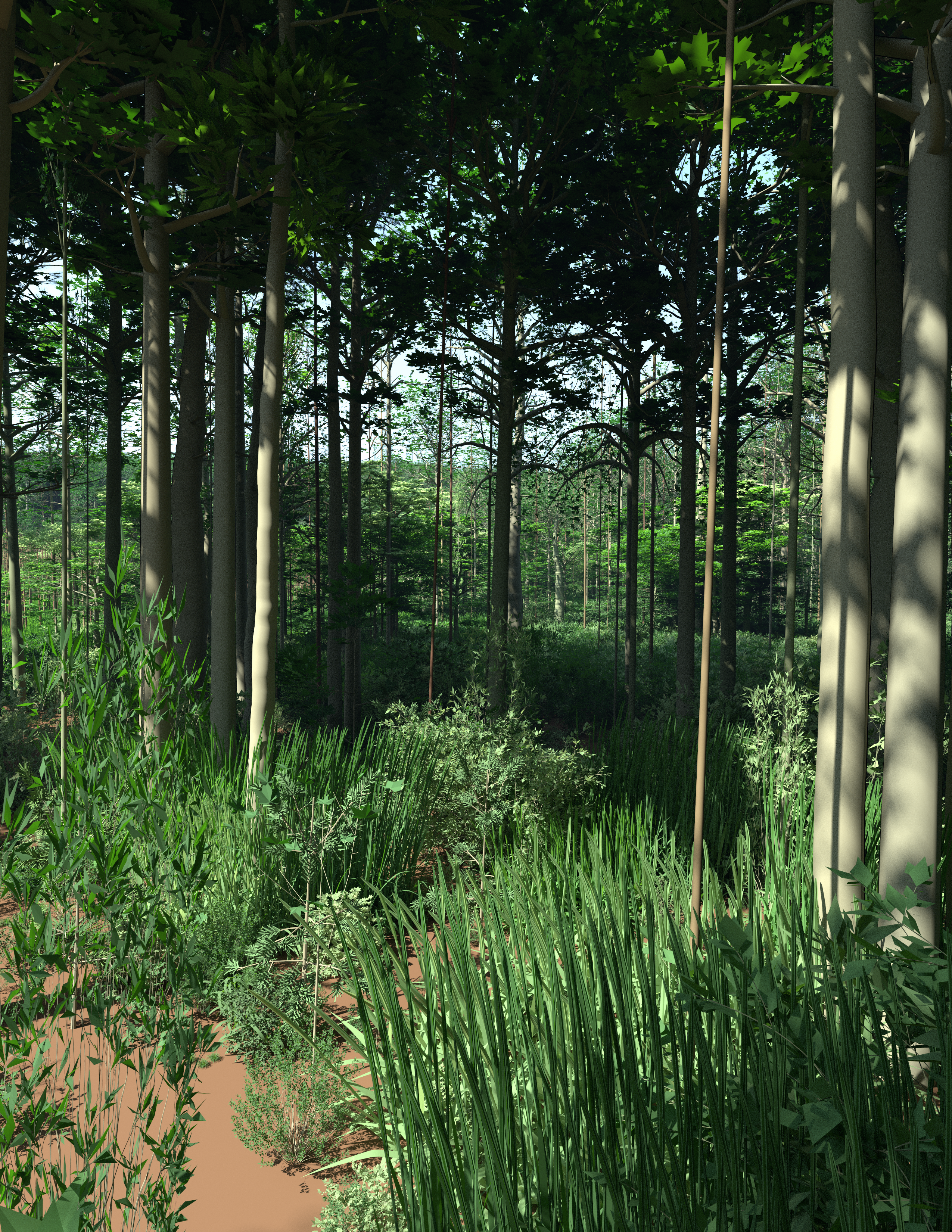} }}
    \subfloat[\centering]{{\includegraphics[width=0.25\textwidth]{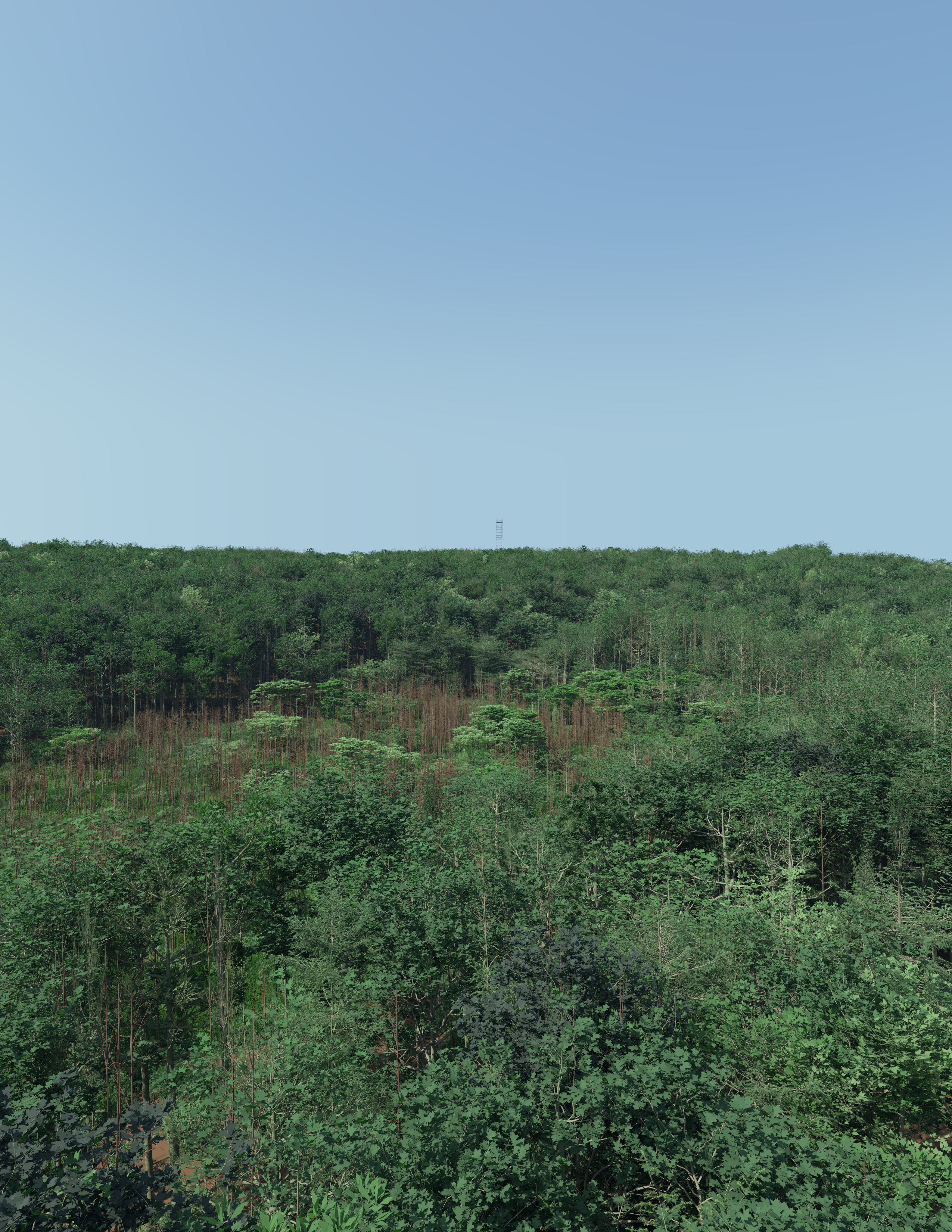} }}
    \caption{Harvard forest scene, simulated using DIRSIG. The simulated scene includes various types of vegetation and objects. (a) is the aerial view, (b) is a within-forest snapshot, and (c) is an above-canopy perspective.}%
    \label{fig:methods:scene}%
\end{figure*}

\section{\textbf{Methods}}

\subsection{\textbf{Study Area}}

\begin{figure}[b!]
    \centering
    \includegraphics[width=0.45\textwidth]{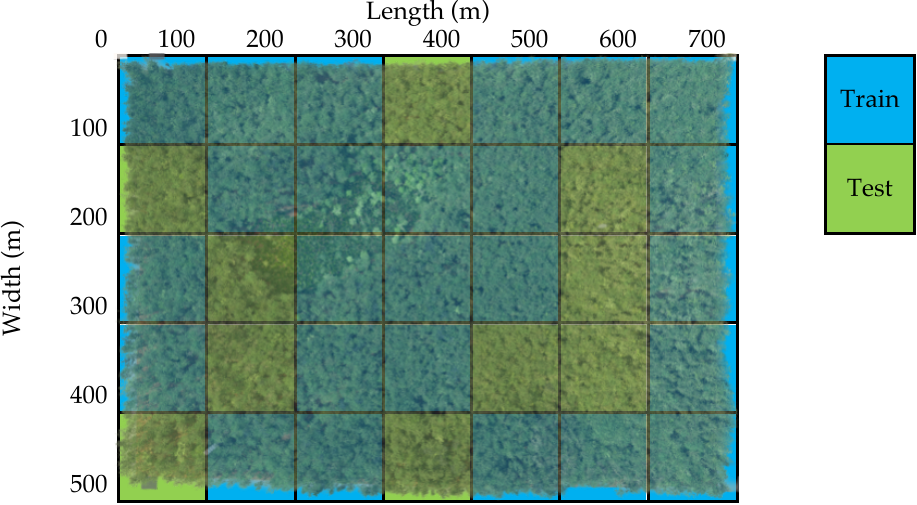}
    \caption{Harvard Forest Scene and the training/validation and testing sets.}
    \label{fig:methods:design_plot}
\end{figure}

In this study, we develop a DIRSIG simulated scene from the Harvard Forest site in Petersham, MA $(42^\circ 32' 19.79''\mathrm{N},\ 72^\circ 10' 31.81''\mathrm{W})$. The Harvard Forest simulated scene is an approximately $500\times700$ meter area consisting of vegetation of different types, such as bushes, grass, trees, as well as soil, and miscellaneous objects, including towers, cars, and tents. Figure~\ref{fig:methods:scene} shows the Harvard forest scene from an aerial perspective as well as within-scene details, and an over-the-canopy perspective. More detailed information regarding the scene can be found in~\cite{chaity2024exploring}. We split the $500\times700$ meter scene into 35 $100\times100$ meter sections, allowing 25 sections to fall under the training and validation set and 10 sections under the testing set (see Figure~\ref{fig:methods:design_plot}). This approach allows for an approximately 27\% testing set. The scene is strategically partitioned into training/validation and testing sets, as some of the targets are underrepresented. 


\subsection{\textbf{RIT VLP-16 Simulated Platform}}
The VLP-16 platform has shown versatility in many areas of research, including remote sensing. The RIT's MX-1 drone suit houses this system, and it has been researched extensively~\cite{Zhangevaluation2022,patki2021assessing,wible2021toward}. We design the simulated VLP-16 platform (RIT VLP-16) in DIRSIG with a transmitter featuring an Elliptical Super Gaussian transverse beam shape. Compared to the real-world VLP-16 setup that includes 16 elliptical beams, the RIT VLP-16 utilizes a single large beam combined with 16 detectors to streamline the configuration. Spectrally, the sensor follows a Gaussian shape centered at 903 nm. Pulses are of Gaussian profile with an energy of 2e-07 J. We design a receiver to process reflected pulses within a defined gate window, covering a distance range of up to 119.8 meters. The RIT VLP-16 also integrates a clock system to coordinate pulse emission and reception, with a pulse rate of 18,080 Hz. Simulating a drone flight, we simulate the platform flying at an altitude of 88 meters with a velocity of 5 m/s, encompassing 4 flight lines while accounting for a maximum 50-meter side overlap in the collected data.


\subsection{\textbf{Simulated Data}}\label{sec:method:data}

To understand the data collection approach in DIRSIG, we must break the workflow into two main steps: First, ``Scene Voxelization" which converts the scene into a 3D grid (voxels) to establish the ground truth, measuring the normalized area for each material. Second, ``Data Voxelization" that transforms the collected RIT VLP-16 data into a similar 3D grid. After both processes, we map and combine the outputs of these two steps, yielding rich ground-truth data on material areas along with spatial and radiometric features - voxel coordinates and signal intensity.


Elaborating further on the scene voxelization step, we first voxelize the Harvard forest scene through a workflow that implements a well-defined, geometric ground-truth calculation for per-material area. The Harvard forest scene is represented as many transformed instances of 3D triangle meshes, which model the physical geometry of exemplars of the tree species. Therefore, it is possible to use these triangle meshes to calculate surface areas directly. In order to obtain the total surface area content in a voxel, we sum the areas of all triangles that intersect it. When a triangle fits entirely inside a voxel, the calculation is clear: simply add its area to the voxel value at hand. However, the astute reader may question the case when a triangle intersects multiple voxels. We effectively clip the triangle to the bounds of each voxel, such that each value $v_{ijk}$ is given by the equation

\begin{equation}
    v_{ijk} = \sum_\ell \mathrm{Area}(\mathcal{V}_{ijk}\cap\mathcal{T}_\ell),
\end{equation}

\noindent where $\mathcal{V}_{ijk}$ is the voxel volume, $\mathcal{T}_\ell$ is the $\ell$th triangle surface, and $\ell$ enumerates all triangles. In practice, we sample $n$ quasi-random, uniformly distributed points from the triangle, such that each point contributes $1/n$ times the total area to the voxel it lands inside. We take $n=32r$ where $r$ is the ratio of the triangle area to the voxel footprint area. This adequately distributes the triangle area into the grid. It is important to note that triangles may be included or excluded from any voxelization depending on the materials we seek to isolate. This enables us to accurately partition the voxelized surface area of the scene geometry into distinct material categories.

\begin{figure}[b]
    \centering
    \includegraphics[width=1\linewidth]{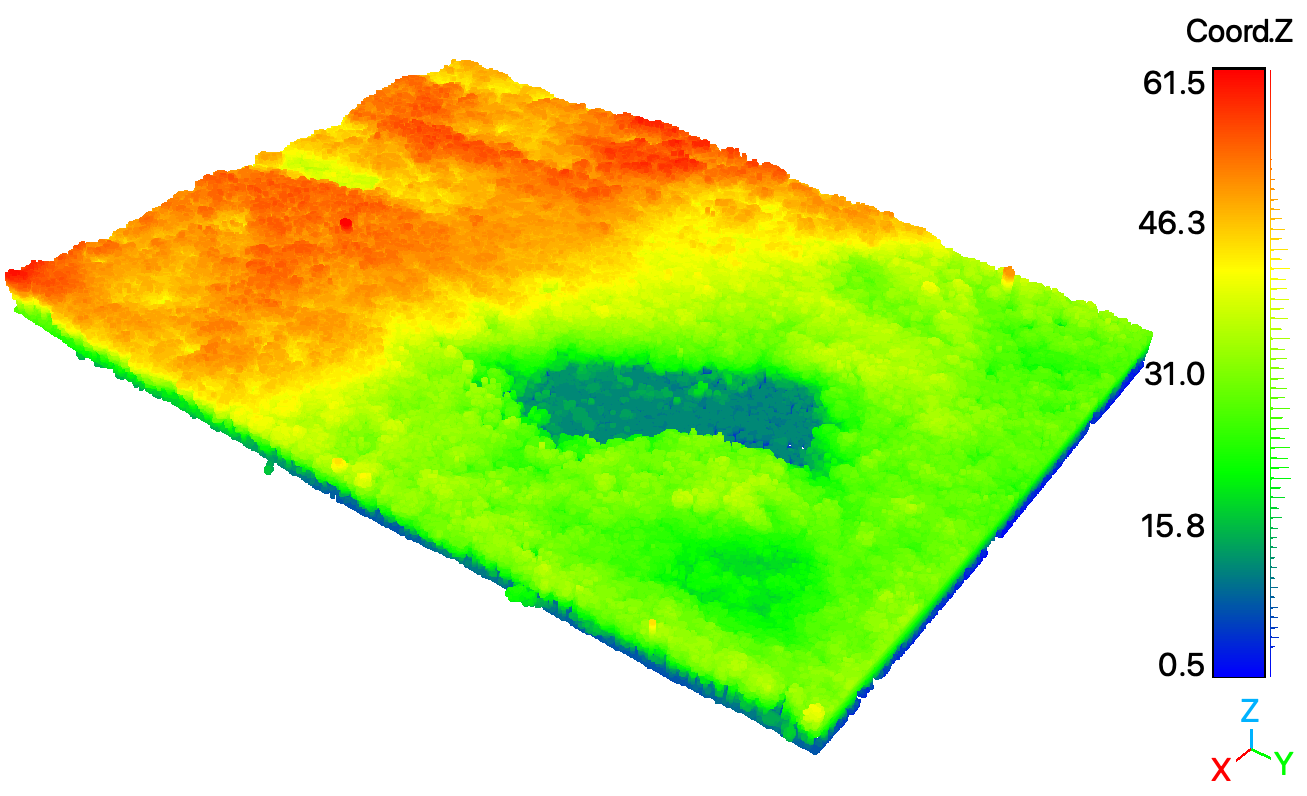}
    \caption{Simulated LiDAR data using DIRSIG software on Harvard forest scene. Color represents height above ground.}
    \label{fig:method:real_data_dirsig}
\end{figure}

Once we complete the DIRSIG simulation, the RIT VLP-16 data is generated in binary format. It is important to note that DIRSIG is a line-of-sight ray-tracing software. Uniquely, instead of tracing a ray from the source to the sensor, it starts from the sensor and is traced back to the source. To that point, we store the collected data in terms of waveforms, representing rays interacting with material. For the RIT VLP-16 setup, we store two returns per waveform. Each set of reconstructed data includes the total scattered fraction of photons, the total remaining fraction of photons, and a number of voxel waveforms. Every waveform is associated with a specific line-of-sight geometric ray, and it could intersect multiple voxels. We assign this ray-interaction data to the corresponding voxels in a predefined grid (the same grid used for calculating area). Subsequently, we calculate the mean scattering density for each voxel, serving as that voxel's signal intensity. The calculations are based on the multiplicative property of the Beer-Lambert law to account for the signal's otherwise diminishing strength as it travels over distance. That is, we use the following equations to estimate the scattered and remaining fractions $f_{\text{scatter}}$ and $f_{\text{remaining}}$, where $t_{\text{near}}$ and $t_{\text{far}}$ are the near and far distances where the ray intersects the voxel, and where $S(t)$ is the waveform signal as a function of distance. Figure~\ref{fig:method:real_data_dirsig} represents the collected simulated data. 

\begin{equation}
    f_{\text{scatter}} = 
    \frac{\int_{t_\text{near}}^{t_\text{far}} S(t)\,dt}{\int_{t_\text{near}}^\infty S(t)\,dt},
\end{equation}

\begin{equation}
    f_{\text{remaining}} =
    \frac{\int_{t_\text{near}}^\infty S(t)\,dt}{\int_0^\infty S(t)\,dt}.
\end{equation}

As mentioned, voxel size plays a crucial role in the performance of the task being utilized. In this study, we approached our objective at various voxel sizes. This would not only ensure a maximized model performance but also identify to what extent the mentioned voxels are representative of the forest structure. We approach this sensitivity analysis by creating voxels of sizes 0.25, 0.5, 1, 1.5, and 2 meters. We depict the datasets generated in this study in Table~\ref{tab:method:data} and their corresponding number of voxels (samples). The scene contains four different materials/targets, namely Bark, Leaf, Soil, and Miscellaneous (Misc). Misc represents cars, towers, and tents, which we put in the forest during the simulation.
\begin{table}[t]
\centering
\caption{Datasets in this study at various voxel sizes and corresponding number of samples. Note that each voxel could contain more than one target. The Harvard forest scenes include four different targets: Bark, Leaf, Soil, and Misc.}

\begin{tabular}{ccc}
\toprule

\textbf{Voxel Size (m)} & \textbf{Partition} & \textbf{\# of Samples (Millions)}\\
\midrule

\multirow{2}{*}{2}   & Training & 0.96  \\
                     & Testing  & 0.26  \\
\midrule

\multirow{2}{*}{1.5} & Training & 1.9     \\
                     & Testing  & 0.53    \\
\midrule

\multirow{2}{*}{1}   & Training & 5.35  \\
                     & Testing  & 1.45  \\
\midrule

\multirow{2}{*}{0.5} & Training & 26.2  \\
                     & Testing  & 7.03  \\
\midrule

\multirow{2}{*}{0.25}& Training & 113    \\
                     & Testing  & 30.4  \\
\bottomrule
\end{tabular}
\label{tab:method:data}
\end{table}

\subsection{\textbf{Analysis}}
\noindent \textbf{Problem Setting.} 
A multi-target regression problem involves predicting multiple target variables simultaneously using a set of input features~\cite{Xu:2019}. Suppose we have a dataset with \(N\) data samples, \(M\) input features, and \(K\) target variables. Let us denote the dataset as \((I, O)\), where \(I\) is an \(N \times M\) matrix representing the input features for $N$ data points:

  \[
  I = \begin{bmatrix}
  i_{11} & i_{12} & \cdots & i_{1M} \\
  i_{21} & i_{22} & \cdots & i_{2M} \\
  \vdots & \vdots & \ddots & \vdots \\
  i_{N1} & i_{N2} & \cdots & i_{NM}
  \end{bmatrix},
  \]
\noindent and \(O\) is an \(N \times K\) matrix representing the target variables for $N$ data point:
  \[
  O = \begin{bmatrix}
  o_{11} & o_{12} & \cdots & o_{1K} \\
  o_{21} & o_{22} & \cdots & o_{2K} \\
  \vdots & \vdots & \ddots & \vdots \\
  o_{N1} & o_{N2} & \cdots & o_{NK}
  \end{bmatrix},
  \]
We can represent the model as a function \(f\) that takes the input features \(I\) and produces predictions \(\hat{O}\) for the \(K\) target variables:

\[
\hat{O} = f(I),
\]
\noindent where \(\hat{O}\) is an \(N \times K\) matrix of predicted target values. The aim is to find the model parameters that minimize a suitable loss function $\mathcal{L}$ that measures the error between the predicted \(\hat{O}\) and the true \(O\) \cite{Rodriguez:2022}. This can be formulated as an optimization problem:
\[
\text{Minimize }   \mathcal{L}(O, \hat{O})
\]
The choice of the loss function depends on the specific problem and the nature of the target variables. We will delve more into the designed loss function for our problem in the following sections.

\smallskip
\noindent\textbf{Voxel Content Estimation via Multi-target Regression.} 
The effectiveness of Kernel Point Convolution (KPConv) has been widely demonstrated across studies~\cite{de2023siamese, li2021comparison, kochanov2020kprnet}. Drawing inspiration from image-based convolution, KPConv leverages kernel points to define where each kernel weight applies, enabling the learning of local geometric patterns. These convolution weights operate within the Euclidean domain, as highlighted by Thomas \textit{et al.}~\cite{thomaskpconv2019}.

\begin{figure*}[t]
    \centering
    \includegraphics[width=1\textwidth]{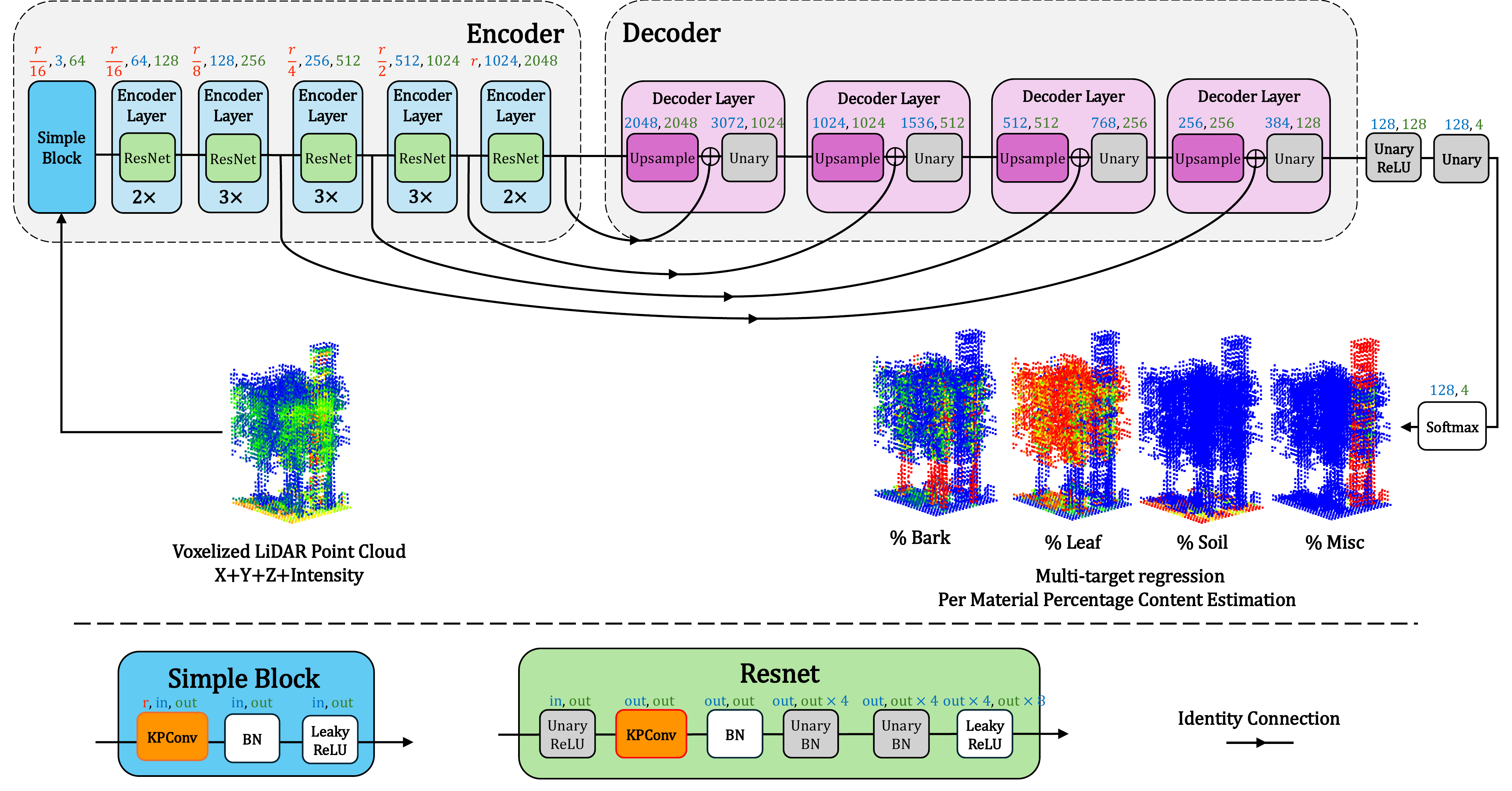}
    \caption{Architecture of KPConv-rigid, consisting of an encoder and a decoder, adapted for multi-target regression. The radius size passed to the kernel point convolution is indicated in red, the number of input features in blue, and the number of output features in green.}
    \label{fig:method:kpconv_model}
\end{figure*}

Fundamentally, these kernel points exert influence based on a correlation function. KPConv's strength lies in its utilization of two distinct sub-sampling strategies—potentials and random sampling—to ensure robustness across varying densities. The deformable variant of KPConv facilitates local shifts in kernel points, enabling adaptation to diverse cloud regions. However, to leverage this adaptability for point cloud classification and segmentation tasks, a regularization step becomes necessary for the deformable convolution to conform to the point cloud geometry. The architectural versatility of KPConv manifests in its rigid and deformable forms. Rigid KPConv demonstrates superior performance in simpler tasks, such as outdoor scenes, while its deformable counterpart outperforms in more complex scenarios with a greater diversity of objects, like indoor scenes. This analysis encompasses both architectures—rigid and deformable—of KPConv. 

We present a visual representation of KPConv rigid's architecture in Figure~\ref{fig:method:kpconv_model}. The KPConv architecture itself comprises an encoder-decoder framework. The encoder consists of multiple layers utilizing kernel point convolutions at various radius sizes to learn features across scales, progressively increasing the number of features learned. The decoder also includes multiple layers, where each layer performs spatial upsampling. Additionally, skip connections pass the features learned from the encoder into the decoder's input, and the combined input is then passed to a unary block, which is essentially a multi-layer perceptron with a single layer and batch normalization. Each layer in the decoder refines the features until the desired output is reached. The architecture takes as input the features \(X, Y, Z\) and intensity, producing an output of \(N \times 4\), where \(N\) is the number of observations, and 4 justifies the four target materials in the multi-target regression task. A Softmax function is applied at the output to ensure that the predictions for each target sum up to 1.

In Figure~\ref{fig:method:kernelpointconvolution}, we illustrate how kernel point convolution works in KPConv. The process follows these steps:

\begin{enumerate}
    \item A sphere of radius \( r \) is defined, centered at point \( x \). Within this sphere, \( H \) neighboring points, denoted as \( x_i \), are identified using the nearest-neighbor algorithm.
    
    \item The neighboring points are centered with respect to \( x \), resulting in \( x_i - x \), which is passed to the influence function \( h \) weighted by $W$ to obtain the kernel \( g \) that is defined as:
    \begin{equation}
            g = \sum h(x_i - x, \tilde{x}_k).W,
    \end{equation}

The function \( h \) calculates the distance between the centered neighboring points \( x_i - x \) and the kernel-defined points \( \tilde{x}_k \) using the \( L_2 \)-norm.
\begin{equation}
        h(x_i - x, \tilde{x}_k) = \max\left(0, 1 - \frac{\|x_i - x - \tilde{x}_k\|}{\sigma}\right),
\end{equation}

where $\sigma$ is the influence distance (see~\cite{thomaskpconv2019} section 3.3). The output of \( h \) is a matrix of size \( H \times K \), and \( W \) is a learnable weight matrix that is randomly initialized and has a size \( K \times C_\text{in} \times C_\text{out} \), where \( C_\text{in} \) is the number of input features, and \( C_\text{out} \) is the number of output features.

\item The resulting matrix from the kernel \( g \) has a size of \( H \times C_\text{in} \times C_\text{out} \). The kernel \( g \) essentially weights the neighboring points based on their proximity to \( \mathbf{x} \) and the kernel-defined points \( \hat{\mathbf{x}}_k \).

\item These weights are used to map the features of the neighboring points \( H \), which have a size of \( H \times C_\text{in} \), into a feature matrix of size \( H \times C_\text{out} \). The final output feature for the point \( \mathbf{x} \) is aggregated from the \( H \) neighboring points, resulting in a vector of size \( C_\text{out} \).
\end{enumerate}

We benchmarked KPConv against three other state-of-the art models: Repsurf~\cite{ran2022surface}, PointNet2~\cite{qi2017pointnet++}, and PointTransformer~\cite{zhao2021point}. The choice of these models was based on their superior performance on the segmentation task.

\noindent \textbf{Loss Function.}
We aim to find the mapping function in Equation~\ref{eq:intro:mapping_functions}, where the input signal to the model consists of voxel coordinates and intensity. We mathematically outline the mapping function as:
\begin{figure*}[t]
    \centering
    \includegraphics[width=1\textwidth]{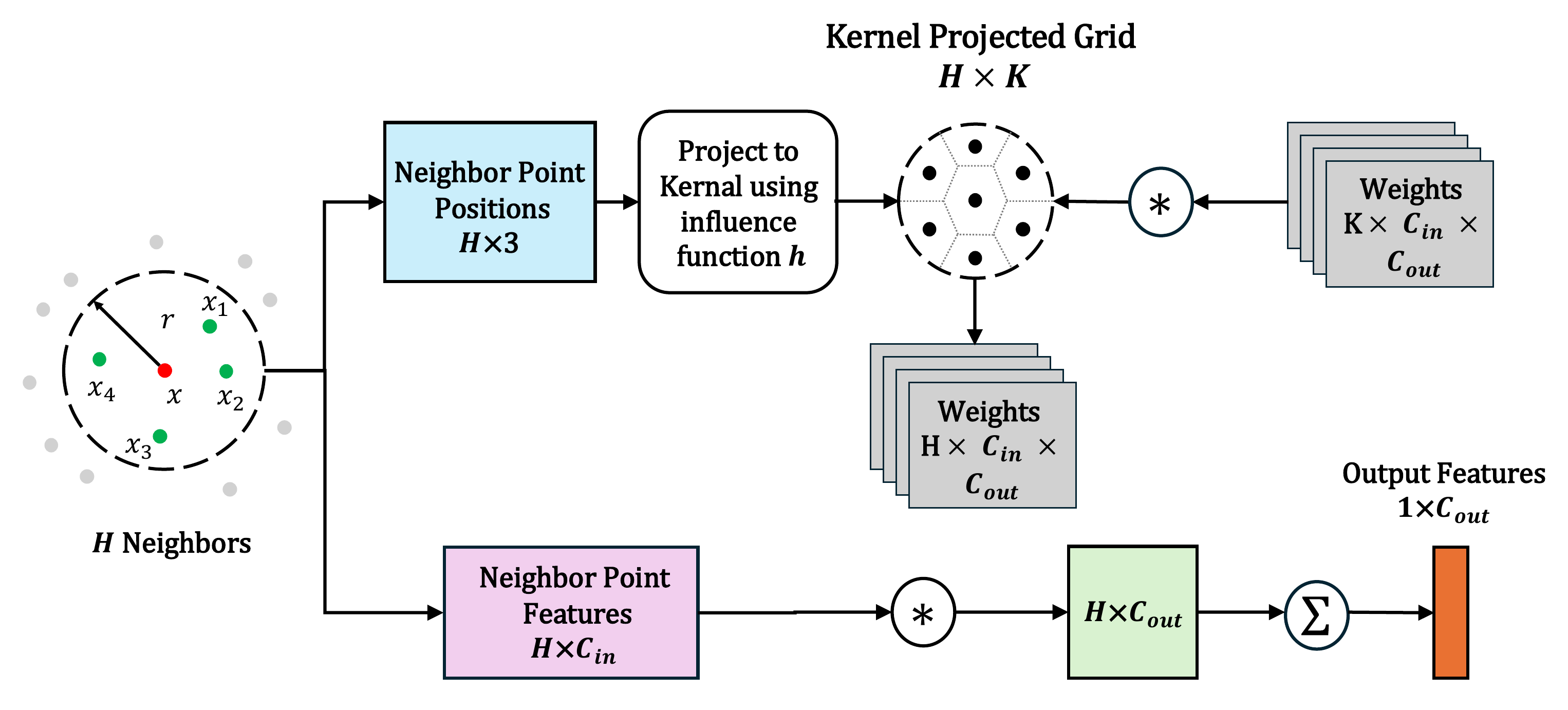}
    \caption{Workflow of kernel point convolution schematic. We adopted and modified the schematic from~\cite{thomas2024kpconvx}.}
    \label{fig:method:kernelpointconvolution}
\end{figure*}
\begin{equation}
f(\text{x, y, z, intensity}) = P_{\text{bark}} + P_{\text{leaf}} + P_{\text{soil}} + P_{\text{misc}}
\label{eq:method:intensity_obj}
\end{equation}
Here, \( P_{\text{target}} \) represents the predicted percentage of a specific target material (e.g., bark, leaf, soil, or Misc) within the voxel. To evaluate the model’s performance, we take Mean Squared Error (MSE), a common regression metric defined as:
\begin{equation}
\text{MSE} = \frac{1}{N} \sum_{j=1}^N (o_j - \hat{o}_j)^2,
\end{equation}
where \( o_j \) is the true value, \( \hat{o}_j \) is the predicted value, and \( N \) is the total number of samples.

To address the class imbalance in the dataset, we integrate a cost-sensitive approach. We define weighted MSE (WMSE) as:
\begin{equation}
    \mathcal{L}_{\text{wmse}} = \frac{1}{N}\frac{\sum_j w_j (o_j - \hat{o}_j)^2}{\sum_j w_j},
\label{eq:method:wmse}
\end{equation}

where \( w_j \) represents the weight assigned to sample \( j \). We derive the weights \( w \) using techniques such as Kernel Density Estimation (KDE), Phi-based relevance, and we porpose Density-Based Relevance (DBR), which we will discuss in detail in the next subsection.

In addition, we capture sample hardness using FocalR, introduced in \cite{yang2021delving} as 
\begin{equation}
    \mathcal{L}_{\text{focalr}} = 
\frac{1}{n} \sum_{i=1}^n \sigma(|\beta e_i|) \gamma e_i
\label{eq:method:focalr}
\end{equation}

where $e$ is the error (MSE), $\beta$ and $\gamma$ are hyperparameters, and $\sigma$ is the Sigmoid function. We also adopt KPConv’s original point-to-point regularizer, denoted as \( \mathcal{L}_{\text{reg}} \). Finally, we define the loss function for a specific target as:

\begin{equation}
    \mathcal{L}_{\text{target}} = \mathcal{L}_{\text{wmse}} + \mathcal{L}_{\text{reg}} + \mathcal{L}_{\text{focalr}}.
\label{eq:method:target_loss}
\end{equation}

We then formulate the total loss function for all targets  as:
\begin{equation}
    \mathcal{L} = \sum\mathcal{L}_{\text{targets}} =  \mathcal{L}_{\text{bark}} + \mathcal{L}_{\text{leaf}} + \mathcal{L}_{\text{soil}} + \mathcal{L}_{\text{misc}}
\label{eq:method:final_loss}
\end{equation}

In this analysis, we explore how different variations of the loss function impact the model performance.

\smallskip

\noindent \textbf{Density Based Relevance.}
Among cost-sensitive approaches in the literature for addressing dataset imbalance is Kernel Density Estimation (KDE;~\cite{steininger2021density}). KDE is a nonparametric approach that estimates the probability density function (PDF) of a random variable, but it relies on a bandwidth parameter. In our analysis, we use KDE and Scott’s bandwidth selection method to determine the bandwidth size~\cite{scott2015multivariate}. Cross-validation can also be used to identify the optimal bandwidth size, but it is computationally expensive, particularly for large datasets such as point clouds. In this study, we normalized the probability density function from KDE and inverted it to serve as a relevance (weighting) measure. Moreover, the SMOTER Phi ($\phi$) is another relevance measure that relies on user-defined parameters. In our analysis, we used SMOTER $\phi$ relevance measure using default parameter values.

\begin{figure}[b]
    \centering
    \includegraphics[width=0.45\textwidth]{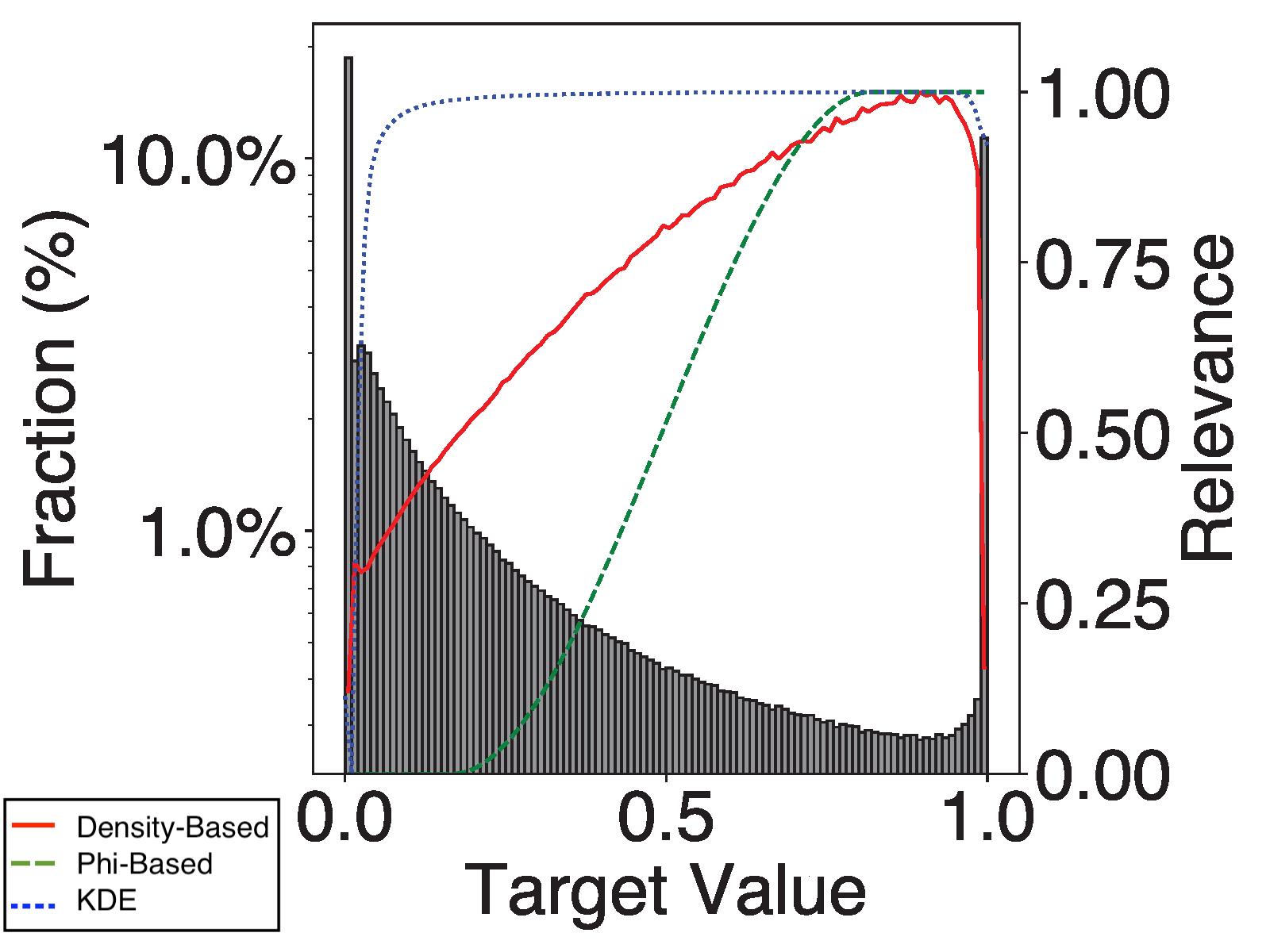}
    \caption{Histogram of Bark target variable for 1m voxel size data set.}
    \label{fig:method:relevance_weights}
\end{figure}

Figure~\ref{fig:method:relevance_weights} illustrates the relevance weights for the Phi-based approach, KDE, and the introduced density-based relevance approach. Neither KDE nor $\phi$ relevance can effectively capture the bimodal distribution of the random variable for the bark target. The $\phi$ relevance measure sets rare samples to zero, while KDE assigns a value of 1 to the most common samples. The density-based relevance approach (we denote as $\gamma$) provides a simple yet effective weighting mechanism based on histograms. $\gamma$ considers both the sample size in each bin and the deviation of the bin from a uniform distribution. This weighting system enables cost-sensitive training through the loss function and can also serve as a balanced metric for performance comparison. To elaborate, the density values $d$ are calculated from the random variable's one-dimensional histogram, and we define the bin density fractions $f$ as:

\begin{equation}
f_{j} = \frac{d_j}{\sum_j d_j} \times 100
\end{equation}

If we assume a random variable with a uniform distribution divided into $b$ bins, each bin $j$ would ideally contain $\frac{100}{b}\%$ of the data. Using this as the uniform scenario for balanced data, we calculate the offset from the uniform distribution by assigning inverse weights based on the fractions $f_j$. Specifically, if $f_j < \frac{100}{b}\%$, the bin receives a higher weight, and vice versa. This is expressed as:

\begin{equation}
w_j = \frac{100/b}{f_j}
\end{equation}

Finally, we normalize the weights and take the square root to exaggerate the effects:

\begin{equation}
\gamma_j = \left(\frac{w_j}{\max(w)}\right)^{1/2}
\end{equation}

\noindent \textbf{Evaluation Metrics.} 
Appropriate evaluation metrics are crucial in assessing models' performance while addressing data imbalance. When evaluating performance on an imbalanced dataset, either the data itself needs to be balanced and typical metrics could be used, or imbalance metrics for the task at hand are to be leveraged. A common approach to balanced regression metrics is the use of weighted metrics, where weights can be calculated using various methods (e.g., $\phi$-based). In other words, the weights employed during cost-sensitive training can also be utilized for imbalance metrics. However, as shown in Figure~\ref{fig:method:relevance_weights} and discussed previously, these methods may not always provide an accurate representation of the data, as exemplified by the behavior of KDE with a bimodal distribution. To address these limitations, we opt to simply report variations of the Mean Absolute Error (MAE) to evaluate performance across distinct data regions. These include:
\begin{itemize}
    \item Sparse MAE ($MAE_s$): MAE for samples from bins with less than 1\% density.
    \item Moderate MAE ($MAE_m$):MAE from bins with densities between 1\% and 5\%.
    \item Dense MAE ($MAE_d$): MAE for samples from bins with densities above 5\%.
    \item Standard MAE: MAE for the entire dataset (balanced assumption).
\end{itemize}

This approach is especially useful for targets such as Misc, where a significant proportion of samples have values of 0\%, providing valuable insights into how the model handles these challenging scenarios.




Lastly, to facilitate easier visualization of model performance across different regions of the target variables, we also chose to display the mean error per bin for each target. This combined approach allows us to qualitatively and quantitatively evaluate the performance across different tests.

%



\begin{figure*}[t]
    \centering
    \includegraphics[width=1\textwidth]{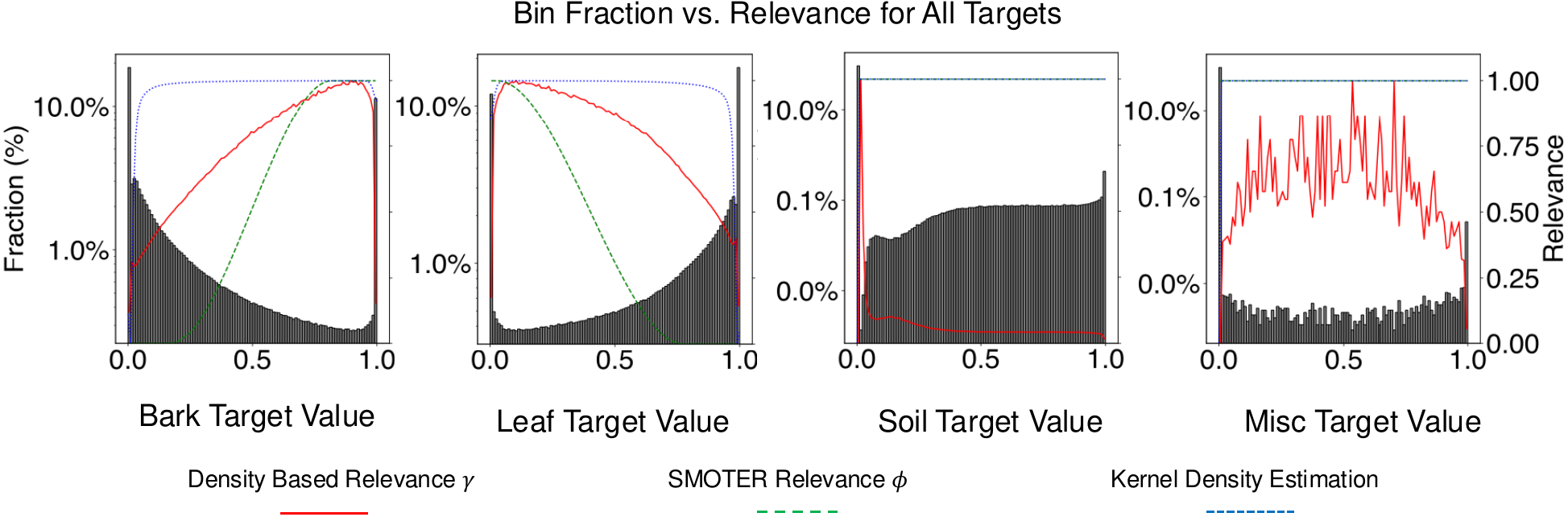}
    \caption{Histogram of the target values and the three weighting approaches used for cost-sensitive learning on 1m voxel size data.}
    \label{fig:results:histogram_of_data}
\end{figure*}

\noindent \textbf{Training Configurations.} Before training the RIT voxel data set, we fully train the models on the DALES point cloud segmentation dataset until convergence~\cite{varney2020dales}. DALES data are among a few Aerial Laser Scanner (ALS) data with a high number of points and targets, including buildings, trees, and cars, among others. Once the model is trained, we leverage transfer learning to DIRSIG voxelized point cloud data. 

We achieve a mean intersection over union (IoU) of 83\% and a mean accuracy of 97.3\% on KPConv, similar to those reported by authors~\cite{varney2020dales}. Hyperparameter tuning was performed using cross-validation on the training dataset.

This study leverages various computational resources, including RIT's research computing facilities~\cite {ritresearchcomputing} and a machine equipped with an Intel(R) Xeon(R) Silver 4214 CPU @ 2.20GHz, 8 NVIDIA RTX 3090 GPUs, and 404 GB of RAM. 

We present the results of this study, starting with the performance of all models using a 1-meter voxel dataset. We then explore different voxel sizes using the best-performing model to conduct a sensitivity analysis. Lastly, we interpret results in the discussion section and the ablation study.


\section{\textbf{Results}}\label{results}
Before delving into the findings of this study, we present preliminary information about the dataset at hand, which will help us better understand the problem and aid the reader in interpreting the following findings with greater ease.

Figure~\ref{fig:results:histogram_of_data} illustrates the distribution of the four targets for 1-meter voxel size training data, along with the three relevance techniques tested in this study: KDE, SMOTER $\phi$ relevance, and $\gamma$ density-based relevance for cost-sensitive training. Starting with the Bark target histogram, approximately 20\% of the data consists of voxels with target values of either 0 or 1; fully Bark (value of 1) or void of Bark (value of 0). The remaining values, ranging between 0 and 1, represent less frequent cases, with their percentages varying from 2\% to less than 0.5\%. The Leaf target histogram exhibits a similar bimodal distribution but in a reversed fashion compared to Bark. This observation aligns with the expected scenario in forest environments, particularly when we view it in voxelized form. For the Soil target histogram, the data primarily reflect voxels that lack Soil, as Soil is typically confined to the ground. Since identifying Soil primarily involves recognizing voxels close to the ground or at the lowest elevation, we expect model performance on Soil targets to be higher than for other variables. Finally, for the Misc target histogram, 99.94\% of the data is void of Misc targets; meaning that a proper performance for this target would be a challenging task.

Due to the kernel fitting method of KDE, the large differences in intensity between bins, and the bimodal nature of the data distribution (particularly pronounced for bark and leaf), KDE’s approach struggles to accurately model the distribution. Specifically, the transitions at the distribution tails are suboptimal, and the relevance values transition abruptly from 0 to 1 when moving between high-density (common samples) and low-density bins (rare). On the other hand, the $\phi$ relevance curve for bark and soil resembles a sigmoid function, with a number of rare bins having no contribution. For soil and Misc categories, $\phi$ assigns a relevance value of 1 to all bins, resulting in overly uniform weighting. In contrast, $\gamma$’s weighting approach for soil might appear sensitive to extremely rare bins. For the misc target, empty voxels are appropriately set to 0, while the remaining bins are adjusted proportionally. 

In the following subsections, we delve into: a) analysis on 1-meter voxel size data, which includes benchmarking against other models and evaluating the effect of different loss functions; and b) voxel size sensitivity analysis across all voxel size datasets using the best-performing model identified in subsection a.

\subsection{\textbf{1 meter Voxel Size Analysis}}

\label{res:1m:model_bench}
\begin{figure}[t]
    \centering
    \includegraphics[width=0.48\textwidth]{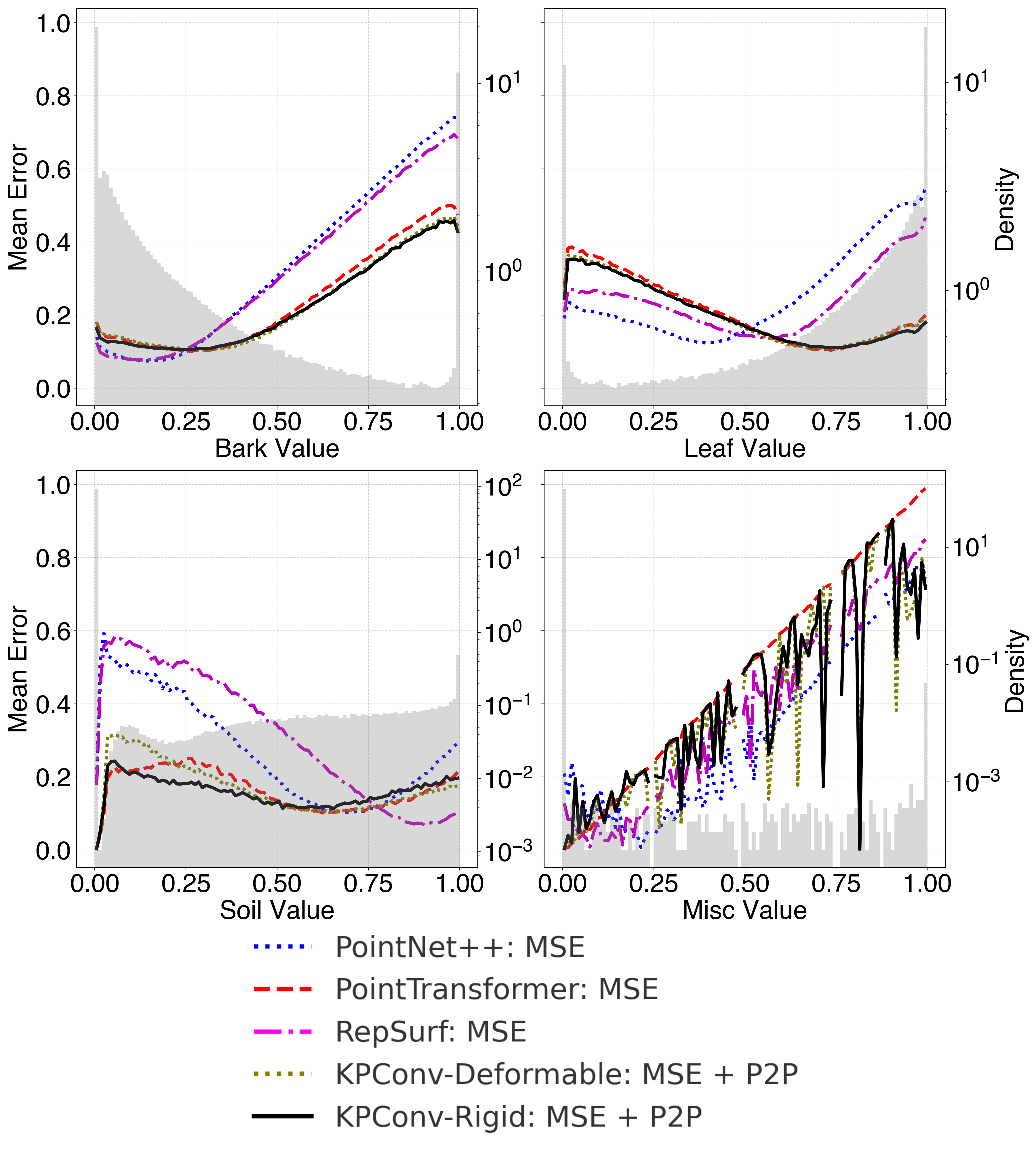}
    \caption{1 meter voxel size model benchmarking qualitative results across four targets. We adopt a simple regression loss here (no cost-sensitive training).}
    \label{fig:results:1m:models_benchmark}
\end{figure}

\smallskip
\noindent \textbf{Model Benchmarking.} We visually present the test partition performance of KPConv in rigid and deformable architectures, PointNet++, PointTransformer, and Repsurf for 1 meter voxel size in terms of per bin mean error (bins = 100) vs. target value; Figure~\ref{fig:results:1m:models_benchmark}. Note that we are not implementing cost-sensitive training in here. We also report density on the log scale on the second y-axis to demonstrate the per-bin density.

At first glance, the figure highlights a clear pattern across models, particularly for the bark and leaf targets. Notably, models with lower errors for low bark values tend to exhibit higher errors for high leaf values, which is due to the two data distributions being mirrors of one another. Overall, KPConv stands out as the top performer, especially in handling rare data. For the soil target, KPConv architectures and RepSurf achieve mean errors that are nearly half of those of other models. As density increases, PointNet++ and RepSurf show improved performances, likely due to sampling techniques designed to process point cloud data that favor denser areas~\cite{qi2017pointnet++,ran2022surface}. Performance for the Misc target is more variable because it is extremely rare, with over 99\% of the voxels without the presence of the Misc target. Interestingly, we can attribute the increasing trend in error for the Misc target to the model interpolating between voxels with Misc values of 0 and 1.

Table~\ref{table:results:1m_modelbenchmarking} reports the quantitative results of MAE across data regions. The sparse MAE ($MAE_s$) is generally higher than that of the moderate and dense regions, which is expected. When comparing models, KPConv architectures (Deformable and Rigid) consistently achieve lower MAE for sparse samples, particularly for bark, leaf, and soil. However, this trend does not hold for Misc targets, where the targets are extremely rare, especially at the 1-meter voxel size; $99.29\%$ of voxels are void of Misc. One could say that due to high values of $MAE_s = 0.46$ for KPConv-Rigid, the models are incapable of accurately regressing values for the misc target, or the predictions for the Misc class is at best random. In the next subsection, an analysis across different voxel sizes with increased sample size would help us understand whether this is the case across all voxel sizes for the Misc target. Findings from this table align with the qualitative visual representations, validating our observations. Based on both qualitative and quantitative assessments, we chose to proceed with the KPConv Rigid architecture for the remaining analyses in this study, as it showed consistent superior performance across Bark, Soil, and Leaf targets.

One key advantage of using KPConv's deformable design is the ability to let the Kernel Point size (K) vary during training, allowing it to adapt to complex shapes and patterns. However, our results show that the rigid design slightly outperformed that of the deformable one, consistent with results from the original study~\cite{thomaskpconv2019}. The effectiveness of a learnable K parameter appears to be dataset-dependent; it thrives in more diverse indoor settings, as corroborated by other studies~\cite{du2022mv}.

\begin{table}[b]
\caption{1 meter voxel size model benchmarking quantitative results. $MAE_s$ represents the mean absolute error for sparse bins (density $<$ 1\%), $MAE_m$ for moderate bins (density 1-5\%), and $MAE_d$ for dense bins (density $>$ 5\%). The models are PN2 (PointNet++), PTv1 (Point Transformer), RS (RepSurf), KP-D (KPConv-Deformable), and KP-R (KPConv-Rigid). Green is the best-performing result, and blue is the second-best.}
\includegraphics[width=0.48\textwidth]{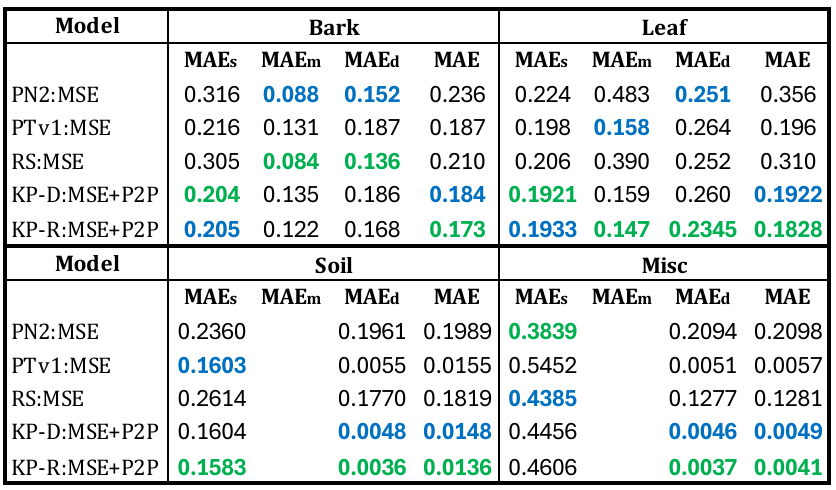}
\label{table:results:1m_modelbenchmarking}
\end{table}

\smallskip
\noindent\textbf{Impact of Cost Sensitive Training.} Previously, we demonstrated that the KPConv-Rigid architecture outperformed other models in benchmarking tests. Based on these results, we select KPConv-Rigid for further analysis. Here, we present results on the impact of different cost-sensitive approaches, specifically $\phi$, $\gamma$, and KDE. Additionally, we examine the effect of implementing FocalR loss for incorporating error regarding harder-to-learn samples~\cite{yang2021delving}.

\begin{figure}[t]
    \centering
\includegraphics[width=0.48\textwidth]{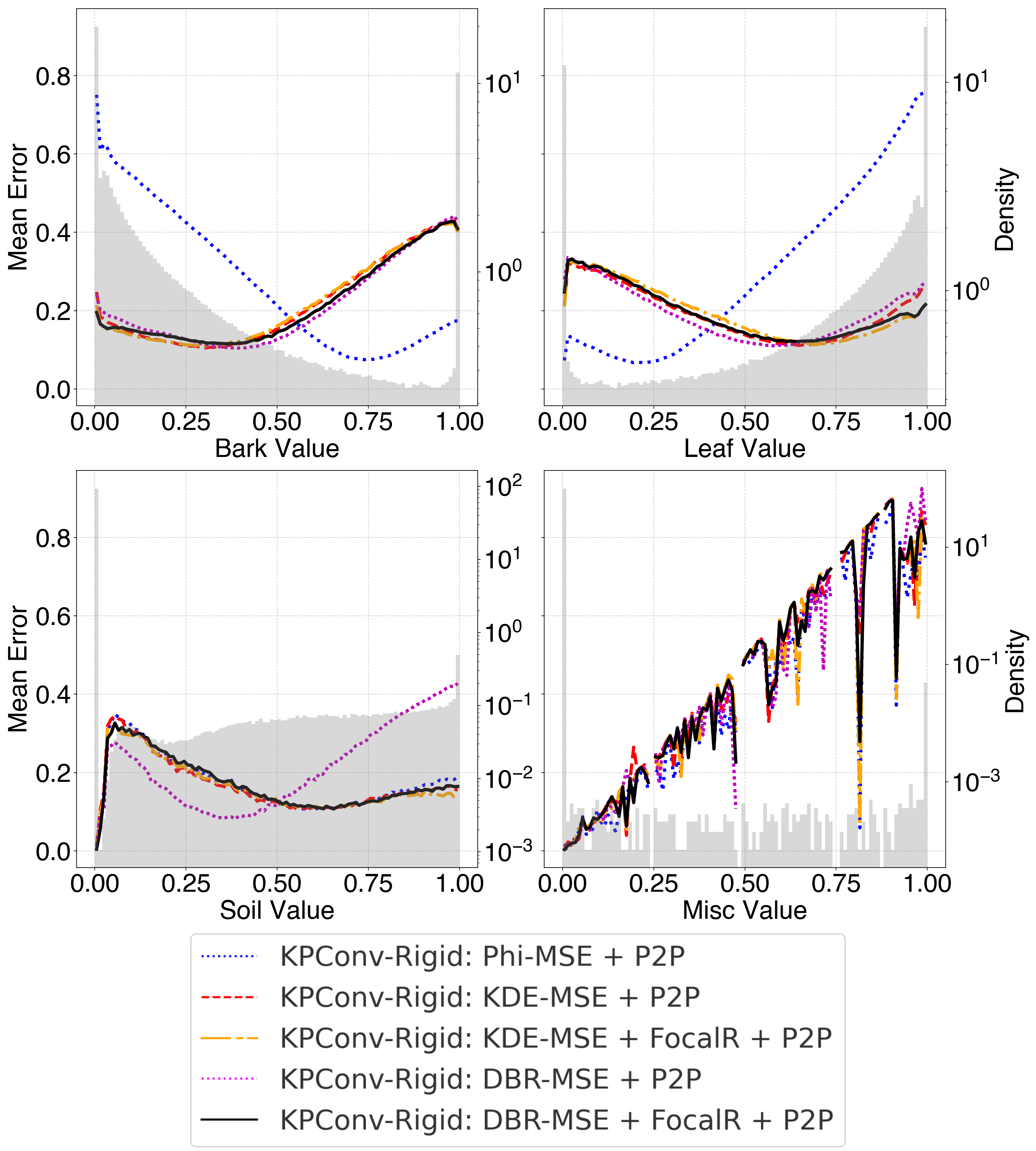}
    \caption{1 meter voxel size cost-sensitive qualitative results across four targets~\ref{res:1m:model_bench}. }
    \label{fig:results:1m:cost_sensitive}
\end{figure}

In Figure~\ref{fig:results:1m:cost_sensitive}, we observe that the performance for rare bins/samples of bark and leaf via $\phi$-based relevance shows significantly lower errors compared to other approaches. This behavior is expected since the $\phi$ relevance value sets many of the weights for bark samples below 0.5 and leaf samples above 0.5 to zero. This causes fewer gradients from the rare samples, which in turn affects the global minima found. For the soil target, the performance across different approaches is generally consistent, except for $DBR\cdot MSE + P2P$. This discrepancy occurs because both $KDE$ and $\phi$ set the densest bin (the first bin for the soil target value) to a weight of zero, assigning all other bins a weight of one. However, in the $DBR\cdot MSE + P2P$ approach, due to the low sample number in a specific bin, the normalization process skews the weighting approach, assigning one bin a disproportionately high weight while lowering the weights of other bins. This highlights a potential drawback of relying solely on the $DBR\cdot MSE + P2P$ approach as a cost-sensitive method. One potential solution is to merge bins with densities below a certain threshold into the previous or next bin, which would help provide a smoother representative distribution of the data. 

That being said, the inclusion of FocalR loss improved the performance of $KDE$ and $DBR$, demonstrating that adding FocalR loss to any cost-sensitive approach enhances generalization and test dataset performance. Specifically, the improvement in $MAE_d$ across four targets shows that FocalR helps the models better identify dense voxels, which are either empty of targets or completely filled with them. FocalR loss also appears to address some of the challenges inherent in DBR normalization mentioned previously. Compared to alternative methods like instance hardness or dynamic instance hardness, FocalR loss may offer an effective yet simple-to-implement solution~\cite{zhou2020curriculum}. For the Misc target, we noted that regardless of the approach used, the magnitude of errors remains high, yielding uninterpretable performance for this target. At the 1-meter voxel size, only 0.07\% of the voxels contain Misc values, making these samples exceedingly rare. While FocalR loss does help to some extent, it is insufficient to extract meaningful patterns from the dataset due to the extremely small sample size for this target. 

The quantitative results in Table~\ref{table:results:1m_costsensitive} align with the qualitative findings, showing that the inclusion of FocalR significantly reduces error for moderate and dense bins across all four targets; 20\% decrease in $MAE_d$ for Bark, approximately $3$ times reduction in $MAE_d$ for Soil, approximately $5-10$ times reduction in $MAE_d$ for Misc. While focal R has a limited impact on sparse bins, it improves overall mean absolute error as well. Notably, for sparse bins, the soil target exhibits the highest mean absolute error due to ineffective DBR weighting caused by normalization; however, FocalR compensates for this issue.

\begin{table}[b]
\caption{1 meter voxel size cost-sensitive quantitative results. The KPConv-Rigid model was used, and different loss functions were compared. Note the decrease in error for $MAE_d$ when including FocalR in the loss function. The best-performing result is highlighted in green, while the second-best result is highlighted in blue.}
\includegraphics[width=0.48\textwidth]{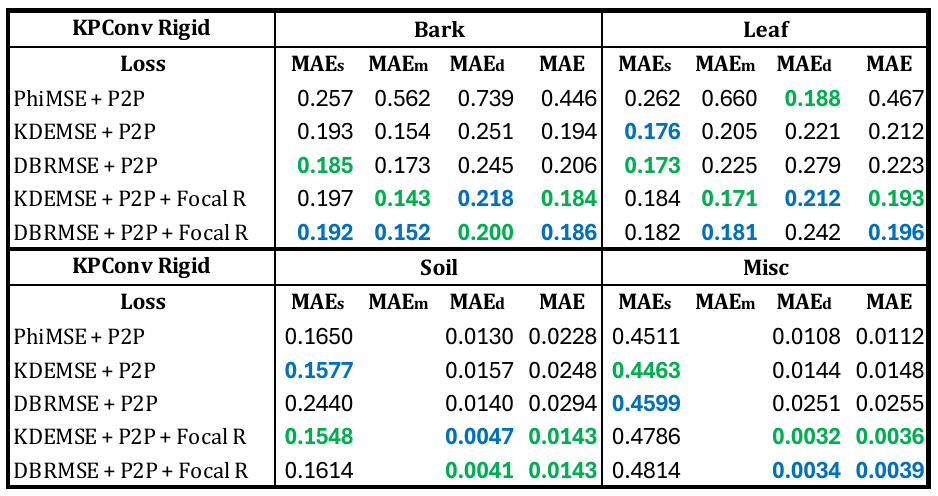}
\label{table:results:1m_costsensitive}
\end{table}

\subsection{\textbf{Voxel Size Sensitivity Analysis}}

\begin{figure*}[t]
    \centering
    \includegraphics[width=1\textwidth]{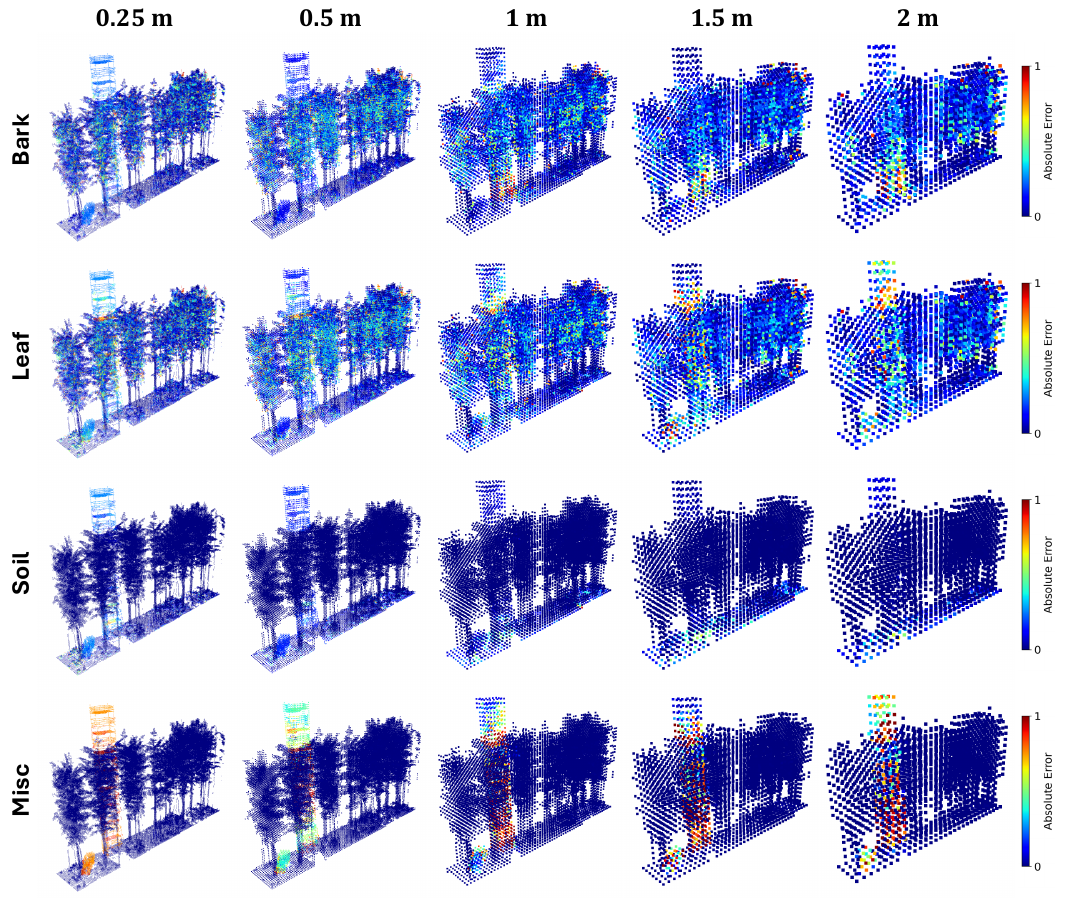}
    \caption{Voxelized clouds for various voxel sizes across the four targets in the tower site. Note the reduction in error within the canopy for bark and leaf going from 0.25 meter voxel size to 2 meter voxel size. Note that despite the misc target being extremely rare, the model successfully detects its presence.}
    \label{fig:results:sensitivity_16}
\end{figure*}

\begin{figure*}[t]
    \centering
    \includegraphics[width=1\textwidth]{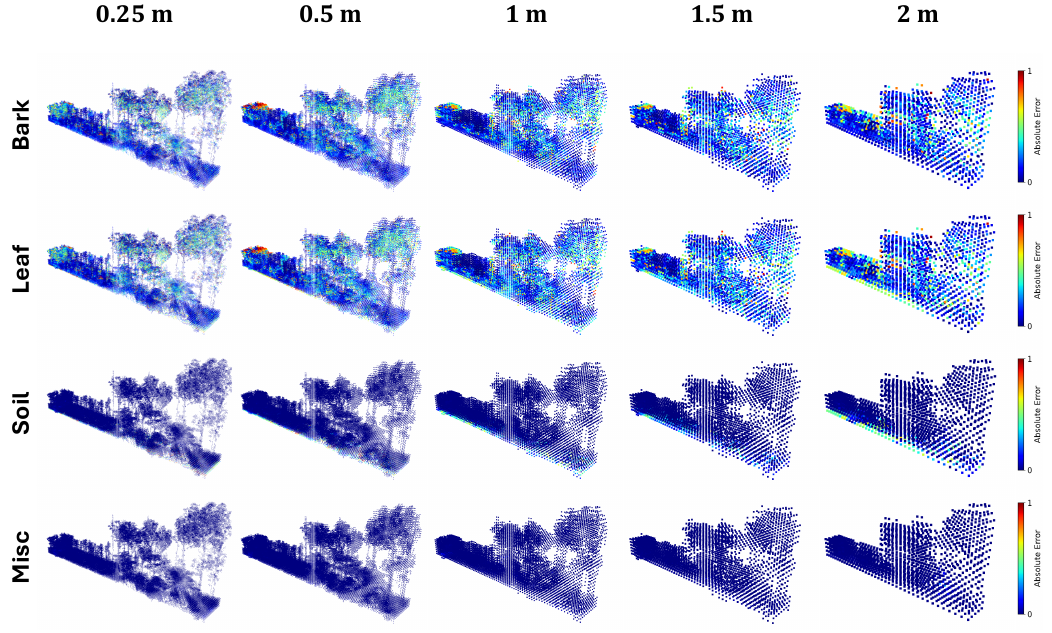}
    \caption{Voxelized cloud for various voxel sizes across the four targets in mud site.}
    \label{fig:results:sensitivity_26}
\end{figure*}

Before delving into the results of the voxel size sensitivity analysis, it is imperative for the reader to keep in mind that the problem at hand is a multi-faceted one. Even though we intend to show the performance results for different voxel sizes, voxel size would not be the only variable when comparing metrics due to the nature of this task. From one voxel size data to another, the size of the testing data, forest structure, and what it represents at a certain voxel size affect the performance metrics. Given that datasets with different voxel sizes have distinct histograms, we chose to present quantitative results (see Table~\ref{table:results:voxelsizesens}) as well as visuals of the absolute error of voxelized clouds (see Figure~\ref{fig:results:sensitivity_16} and ~\ref{fig:results:sensitivity_26}). For this analysis, we choose the KPConv with $DBR\cdot MSE + FocalR + P2P$ loss from the previous section, as it outperformed other loss variations.

\noindent\textbf{Bark and Leaf}. Table~\ref{table:results:voxelsizesens} shows that with an increase in voxel size, we observe lower errors ($MAE_s$ and $MAE_m$) for Bark and Leaf targets. This holds true as within the canopy, there is a large amount of variability incorporated in lower voxel size datasets compared to higher voxel size datasets, thereby yielding higher performance. While this reduction in error is expected, it comes at the cost of losing detailed information about the forest details. This is further validated by the similarities in the locations of errors in Bark and Leaf within the canopy, as in Figure~\ref{fig:results:sensitivity_16} and~\ref{fig:results:sensitivity_26} across different voxel sizes. 

\noindent\textbf{Soil}. Error metric $MAE_s$ for soil across all voxel sizes except 0.25 meter are close, given that soil/terrain should be straightforward to locate and calculate the area for, i.e., a voxel's area is a rough estimate of the terrain area encompassed in the voxel. On the ground truth data side, calculation of the terrain area becomes a more difficult task to do for smaller voxel sizes due to triangulation, which is a potential cause for the sudden increase in $MAE_s$ error for 0.25-meter voxel. At small voxel sizes, bushes and other vegetation introduce more complexity, as a single bush can span multiple voxels, creating greater variability and higher errors. Moreover, the lowest $MAE_s$ is observed at the 1-meter voxel size, indicating an optimal balance between 0.25 meters and the coarser aggregation at 2 meters. 

\noindent\textbf{Misc}.
Misc target shows the lowest performance $MAE_d$ and $MAE$ at 0.25 meter voxel size, primarily because a significant portion of the data contains zero values for Misc compared to higher voxel sizes. $MAE_s$ representing error for sparse sample, at 0.25 meter voxel size, is higher than that of 2 meter voxel size. As the number of empty-of-misc voxels increases, the variability decreases, leading to higher performance. We also notice an increase in error, especially in the proximity of the canopy. Based on the error metric, one can say that the prediction for this variable, given the high mean error for sparse samples, is, at best, random. On the other hand, one can observe from Figure~\ref{fig:results:sensitivity_16} that the model correctly identifies voxels with the presence of the Misc target (from a segmentation perspective). It is also evident that the model exhibits higher error in areas where bark, leaf, and misc targets are adjacent to each other. 

\begin{table}[t]
\caption{Voxel size sensitivity analysis quantitative results. $MAE_s$ is the mean absolute error for sparse bins with density less than 1\%, $MAE_m$ for moderate bins with density 1-5\%, and $MAE_d$ for dense bins with density above 5\%. We used the KPConv-Rigid model with $DBRMSE + FocalR + P2P$ loss.}
\includegraphics[width=0.48\textwidth]{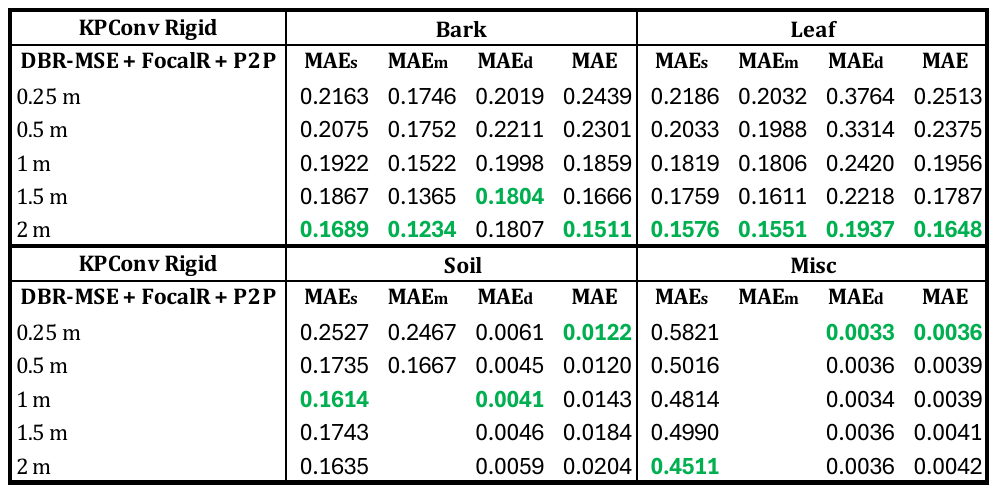}
\label{table:results:voxelsizesens}
\end{table}

\section{\textbf{Discussion}}

We need to keep in mind the multi-faceted nature of the problem while interpreting the findings presented in the previous section - the size of the test set and what data at a specific voxel size represent in terms of the forest, as well as the ability of the model to accurately estimate the target values. The two factors in this analysis, the effect of voxel size on content estimation and the impact of ground truth voxel size data, can be argued. First, regarding the impact of ground truth voxel data, estimating the voxel content at 2 meters is a reasonably simpler task to accomplish than doing so at a 0.25-meter voxel size. The reason is, as mentioned, there is a larger variability in within-canopy measurements at the 0.25-meter voxel set, whereas this is not the case in 2-meter voxel data, and even a rough estimate would provide higher accuracy. The $MAE_s$ and $MAE_m$ metrics for Bark and Leaf at 0.25 and 0.5 meter voxel size represent how difficult a task it is for the model to be able to correctly estimate within-canopy voxel content (Figure~\ref{fig:results:sensitivity_16} and~\ref{fig:results:sensitivity_26}). Additionally, a potential factor for this decrease in performance could be the calculation of the area. Calculating the area for geometrical ground truth at lower voxel sizes becomes a more challenging task due to triangulation. In other words, determining the designated area within a voxel for a target necessitates oversampling, and with larger voxel sizes. Thus, calculating the area is simpler due to the higher number of samples within the original defined triangle. A potential explanation for higher $MAE_s$ and $MAE_m$ errors for Soil at 0.25 meter could be that since Leaf also represents vegetation at ground level (shrubs and bushes), estimating Leaf content closer to the ground would be an easier task to do at a coarser voxel size compared to finer voxel size, again due to the complex structure that vegetation introduces. In short, the choice of voxel size is application-dependent, i.e., it depends on the level of detail needed by the user. Smaller voxel sizes are preferable for retaining detailed information about the forest structure, while larger voxel sizes are better suited for tasks requiring simplified and aggregated data.

\subsection{\textbf{Ablation Studies}}
To systematically evaluate the impact of different model components, we perform an ablation study on the regression metric, kernel size and sphere radius, as well as the speed across the five benchmark models. This helps us determine how individual components contribute to overall performance.

\noindent\textbf{Regression Loss}. 
We perform a test performance comparison between mean squared error (MSE) and L1 distance as regression metrics, evaluating both with and without cost-sensitive training - KDE and DBR on 1-meter voxel size data. The results, shown in Table~\ref{table:ablation:regression_metric}, indicate that L1 consistently outperforms MSE, particularly for sparse targets. This trend is evident across all three variations of mean absolute error. The density-based relevance approach applied to bark and leaf targets, along with L1 as the regression metric, yields the lowest error. One reason why L1 outperforms MSE is that the constant gradient is more stable and less sensitive to large errors compared to L2 distance (MSE). One might also say that MSE tends to overemphasize denser regions, leading to lower performance in terms of generalization. This suggests that future studies could benefit from incorporating L1 as a regression metric.

\begin{figure*}[t]
\includegraphics[width=0.48\textwidth]{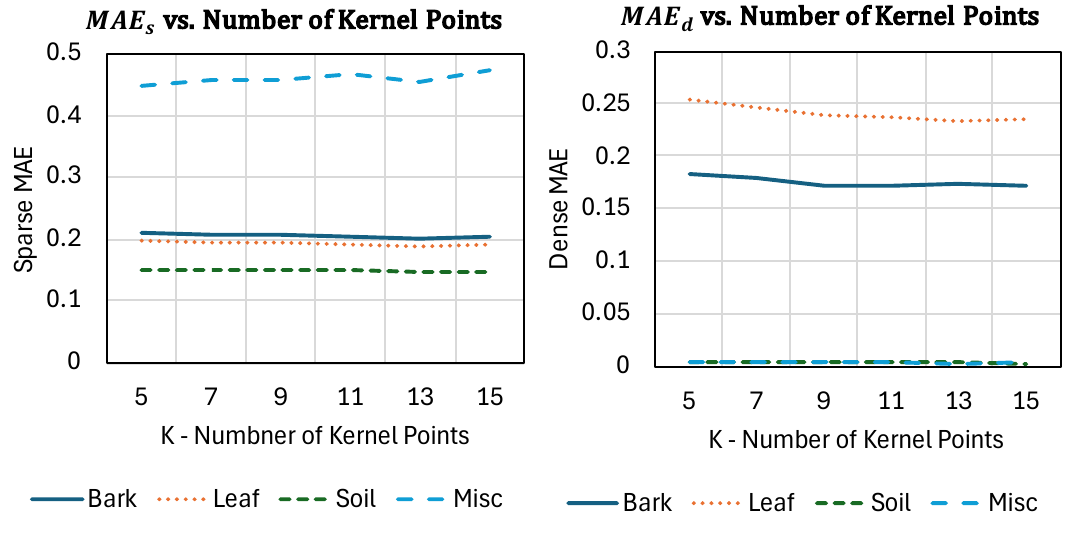}
\includegraphics[width=0.48\textwidth]{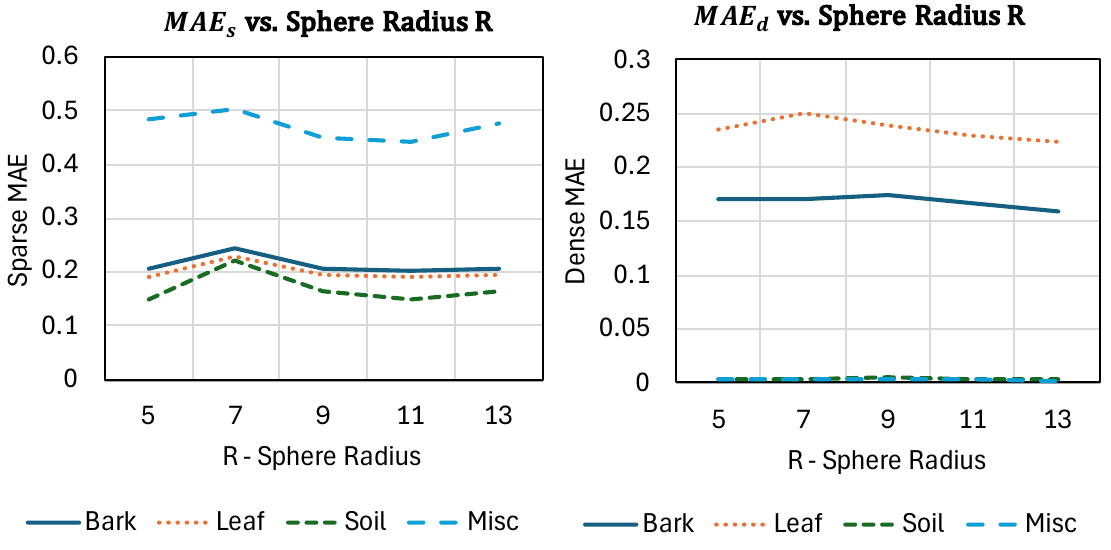}
\caption{Impact of kernel point size (K) and input sphere radius (R) on $MAE_s$ and $MAE_d$ across all four targets for 1 meter voxel data. Note that increasing K and R reduces error.}
\label{figure:ablation:kr}
\end{figure*}

\noindent\textbf{Kernel Point Size and Sphere Radius}. 
Among KPConv’s hyperparameters, two significantly impact performance: the kernel point size (K) and the input sphere radius (R). Here, we evaluate the effects of varying K and R values on 1-meter voxel data using only $MSE$ as loss (no cost-sensitive training), with results presented in Figure~\ref{figure:ablation:kr}. The findings show that errors for $MAE_s$ and $MAE_d$ across all four targets generally decrease as K and R increase, or the changes remain marginal. Additionally, it is important to note that the behavior of K may vary depending on the dataset, which aligns with prior research~\cite{thomaskpconv2019}. We expect that the impact of K and R on performance may differ at various voxel size scales.

\begin{figure}[b]
\centering
\includegraphics[width=0.4\textwidth]{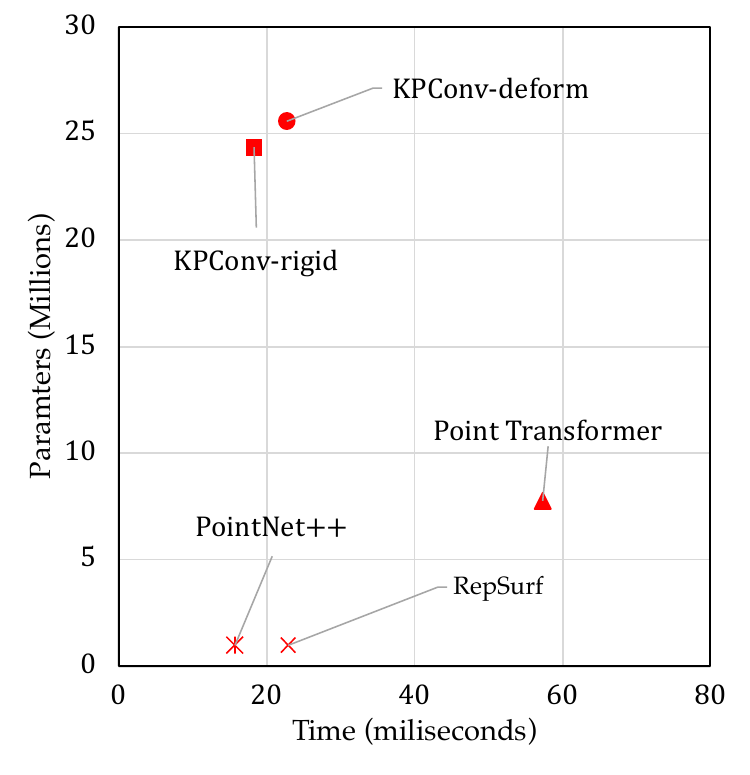}
\caption{Comparison of training time and number of parameters across the five benchmarked models.}
\label{fig:ablation:speed}
\end{figure}

\noindent\textbf{Model Performance and Training Time}. 
We evaluate the performance and the speed of the five benchmark models in this study based on average single batch training and the number of parameters (see Figure~\ref{fig:ablation:speed}). We average the time over 1000 batches, each containing 1000 observations. KPConv architectures are significantly larger than Repsurf and PointNet2 in terms of the number of parameters. It is important to note that Repsurf and PointNet2 have similar parameter counts and inference times because Repsurf builds upon PointNet++ as its foundation model, adding umbrella surface learning. KPConv’s larger size is due to its kernel-based convolutions. However, despite their size, KPConv models exhibit lower inference times compared to the other benchmarked models. We suspect this is due to the overhead associated with distributed data parallelism used in Repsurf, PointNet++, and PointTransformers. Specifically, distributing data across multiple GPUs and gathering results increases processing time, potentially contributing to slower training for these models.

\begin{table}[b]
\caption{Comparison of test performance using L1 and MSE as regression metrics, evaluated with and without cost-sensitive training (KDE and DBR) on 1-meter voxel size data. Note L1 outperforming MSE.}
\includegraphics[width=0.48\textwidth]{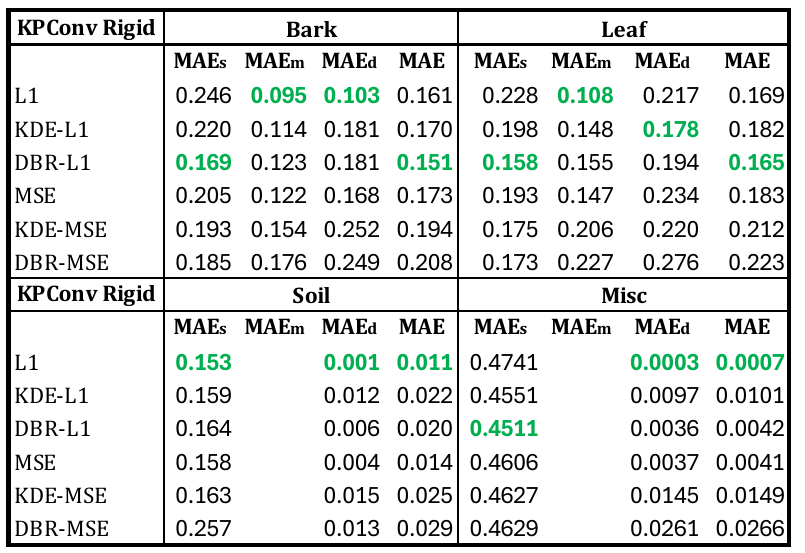}
\label{table:ablation:regression_metric}
\end{table}

\section{\textbf{Conclusions}}\label{conclusion}
LiDAR systems have gained considerable attention due to their ability to capture structural information, which is particularly important in forest environments. A large and growing body of literature focuses on processing LiDAR data in both waveform and discrete forms. One approach to processing LiDAR data is to represent data as voxel grids, reducing the high computational cost. Evidently, the biggest drawback of voxelizing is the loss of information associated with presenting the collected information in a coarser resolution. The question is, can we recover the low-level information that was present prior to voxelization based on the voxelized grid? 

One potential avenue to address this challenge involves modeling the phenomenon; however, this approach is constrained by the availability of ground-truth data. Scene simulation could be used as an alternative to ground truth data collection. In this study, we utilize DIRSIG, a simulation model renowned for its ability to create scenes with accurate radiometric and geometric properties, as well as its capacity to generate a substantial volume of data samples, to address this question. We simulate the Harvard Forest site in Petersham, MA, using DIRSIG. The simulated scene is categorized into Bark, Leaf, Soil, and Misc. One of the advantages of using this simulated scene is the abundance of ground truth data available, which enables the use of supervised deep models. We also generate various voxelized representations at 0.25, 0.5, 1, 1.5, and 2 meters for sensitivity analysis of the forest environment.

Given the potential for each voxel to encompass four distinct targets (Bark, Leaf, Soil, Misc), the task evolves into a multi-target regression, necessitating the prediction of four distinct values. Moreover, to address data imbalance across targets, we implement various cost-sensitive approaches, along with a simple density-based relevance (DBR) method. Our analysis includes (1) benchmarking different deep models, including KPConv, for multi-target regression; (2) evaluating the impact of various loss functions on final performance within a cost-sensitive training framework; and (3) conducting a sensitivity analysis across different voxel sizes.

Our findings highlight the potential for inferring voxel content with acceptable accuracy from voxelized grids within a simulated scene. Benchmarking results reveal that at a 1-meter voxel size, KPConv consistently outperforms other models across all targets. Given its superior performance, we conducted subsequent analyses using KPConv. The evaluation of cost-sensitive approaches demonstrates that both Kernel Density Estimation (KDE) and Density-Based Relevance (DBR) cost-sensitive approaches yield similar performance. Additionally, our investigation of loss functions revealed that combining cost-sensitive training with FocalR loss for hard-to-learn samples yields the best overall performance.

Our sensitivity analysis highlights its substantial impact on model performance. Results depict lower error metrics ($MAE_s$ = 0.168 for Bark and $MAE_s$ = 0.1576 for Leaf) for a 2-meter voxel size. This improvement is associated with reduced variability in the ground truth data, thus over-simplification of the intricate forest structure. Conversely, the 0.25 and 0.5 meter voxel sizes showcase high $MAE_s$ = 0.21 values, indicating diminished model performance for rare samples and accentuating the increased complexity within the forest structure at smaller voxel sizes. Additionally, we note that computational limitations exist when dealing with smaller voxel sizes. The misc target is regarded as extremely rare, and the model performance across different tests is not justified. However, the visualized voxelized clouds in Figure~\ref{fig:results:sensitivity_16} show that the models can successfully detect the presence vs. non-presence of Misc. An optimal voxel size of 1 meter appears to be ideal for soil, showing minimal loss in performance across various voxel sizes. This choice of voxel size proves to be application and user-dependent, with smaller voxel sizes retaining much of the complexity of the forest structure without encountering significant computational limitations, and larger voxel sizes smoothing out the inherent variability. Our work fills the gap in techniques and datasets for multi-target regression in imbalanced datasets.

\noindent\textbf{Future Work.}
On the simulated scene side, future work will focus on addressing additional factors, including point deficiency and the morphological similarity of targets.
Future work will also focus on transferring the learned concepts to real-world datasets. Implementing this in the current study is beyond its scope, as this study is already comprehensive. We plan to use TLS data collected from a forest environment and fine-tune the model to adapt to the real-world data space. Many high-level features in the forest setting remain consistent, so the focus will be on freezing key learned parameters while adjusting for low-level features. For TLS data, the ground truth voxel material percentages will be determined from labeled segmented data. The percentage of each material in a voxel will be calculated based on the proportion of labeled points for that material relative to all labeled points within the voxel. 

For future research, we will aim to leverage DIRSIG for generating simulated 3D scene patches with more uniformly distributed targets across the scene patches. This generates a balanced dataset, which can be used to evaluate the performance of the utilized models and, in turn, compare the performance in both balanced and imbalanced training schemes. On the model side, KPConv’s architecture with an attention mechanism should be studied, as it could aid in identifying important areas of the forest environment and enhance learning. Last but not least, future work will consider further refining DBR cost-sensitive training to better account for extremely rare samples in the dataset.

\section*{Acknowledgment}
The authors would like to thank the Digital Imaging and Remote Sensing Image Generation (DIRSIG) support team, especially Byron Eng, for their contributions regarding simulations. 

This research was supported by an academic grant from the National Geospatial-Intelligence Agency (Award No. HM0476-20-1-000), Project Title: Enhanced 3D Sub-Canopy Mapping via Airborne/Spaceborne Full-Waveform LiDAR. Any opinions, findings, conclusions, or recommendations expressed in this material are those of the author(s) and do not necessarily reflect the views of NGA, DoD, or the US government. Approved for public release,
NGA-U-2025-01661.

\ifCLASSOPTIONcaptionsoff
  \newpage
\fi

\bibliographystyle{ieeetr}
\bibliography{IEEEabrv,references}

%
\begin{IEEEbiography}[{\includegraphics[width=1in,height=1.25in, clip,keepaspectratio]{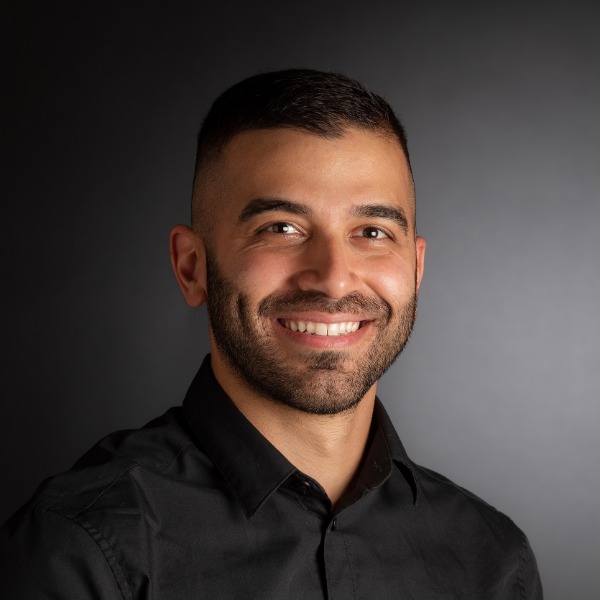}}]{Amirhossein Hassanzadeh} is a Research Engineer in Chester F. Carlson’s Digital Imaging and Remote Sensing (DIRS) lab at Rochester Institute of Technology. He received his B.S. degree in Engineering from Guilan University in Rasht, Iran, in 2016 and a Ph.D. in Imaging Science from Rochester Institute of Technology in 2022. His research interests span across hyperspectral systems, visible and thermal sensing, satellite and drone remote sensing, precision agriculture, and modeling. He is particularly passionate about bridging the gap between remote sensing and AI, aiming to amalgamate these fields for enhanced outcomes.
\end{IEEEbiography}
\begin{IEEEbiography}[{\includegraphics[width=1in,height=1.25in, clip,keepaspectratio]{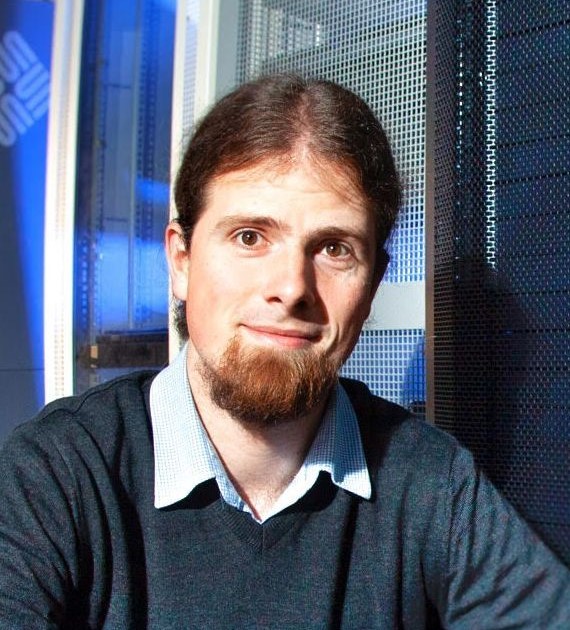}}]{Bartosz Krawczyk} is an assistant professor in the Chester F. Carlson Center for Imaging Science at the Rochester Institute of Technology. He obtained his M.Sc. and Ph.D. degrees in Computer Science from Wroclaw University of Science and Technology, Poland, in 2012 and 2015, respectively. Dr. Krawczyk's current research interests include machine learning, continual and lifelong learning, data streams and concept drift, class imbalance, and explainable artificial intelligence. He has authored more than 70 journal papers and over 100 contributions to conferences. Dr. Krawczyk coauthored the book Learning from Imbalanced Data Sets (Springer 2018). He was a recipient of prestigious awards for his scientific achievements, such as the IEEE Richard Merwin Scholarship, IEEE Outstanding Leadership Award, and Amazon Machine Learning Award and Best Paper Award at CLVISION CVPR Workshop. He served as a Guest Editor for four journal special issues and as a Chair for twenty special sessions and workshops. Dr. Krawczyk is a Program Committee member for high-ranked conferences, such as KDD (Senior PC member), AAAI, IJCAI, ECML-PKDD, ECAI, PAKDD, and IEEE BigData. He is a member of the editorial board for Applied Soft Computing (Elsevier).
\end{IEEEbiography} 
\begin{IEEEbiography}[{\includegraphics[width=1in,height=1.25in,clip,keepaspectratio]{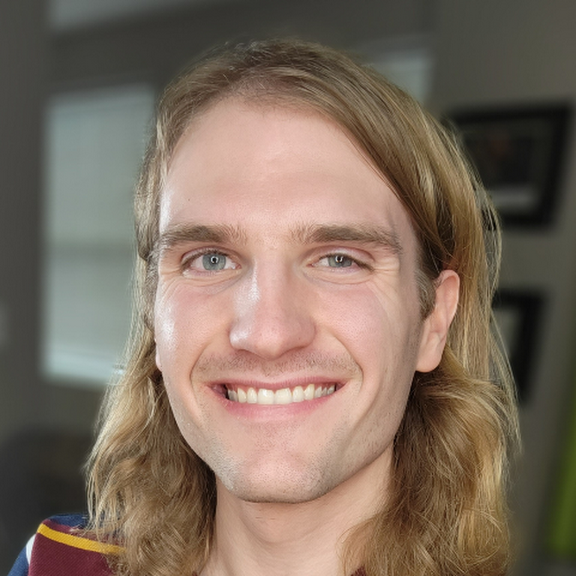}}]{Michael Saunders}
Grady is a research scientist at the Carlson Center for Imaging Science at the Rochester Institute of Technology, having earned his Master's degree from the institution in 2020. His academic journey is rooted in a childhood passion for the visual arts, which evolved into an intense curiosity for science and technology. He is primarily interested in the convergence of traditional computer graphics with the mathematical modeling and simulation of physical, ecological, or otherwise Natural phenomenology. Presently, Grady dedicates the majority of his professional efforts to supporting the advancement of the Digital Imaging and Remote Sensing Image Generation (DIRSIG) software package, actively contributing to its development, and collaborating on research projects that leverage its capabilities.
\end{IEEEbiography}
\begin{IEEEbiography}[{\includegraphics[width=1in,height=1.25in,clip,keepaspectratio]{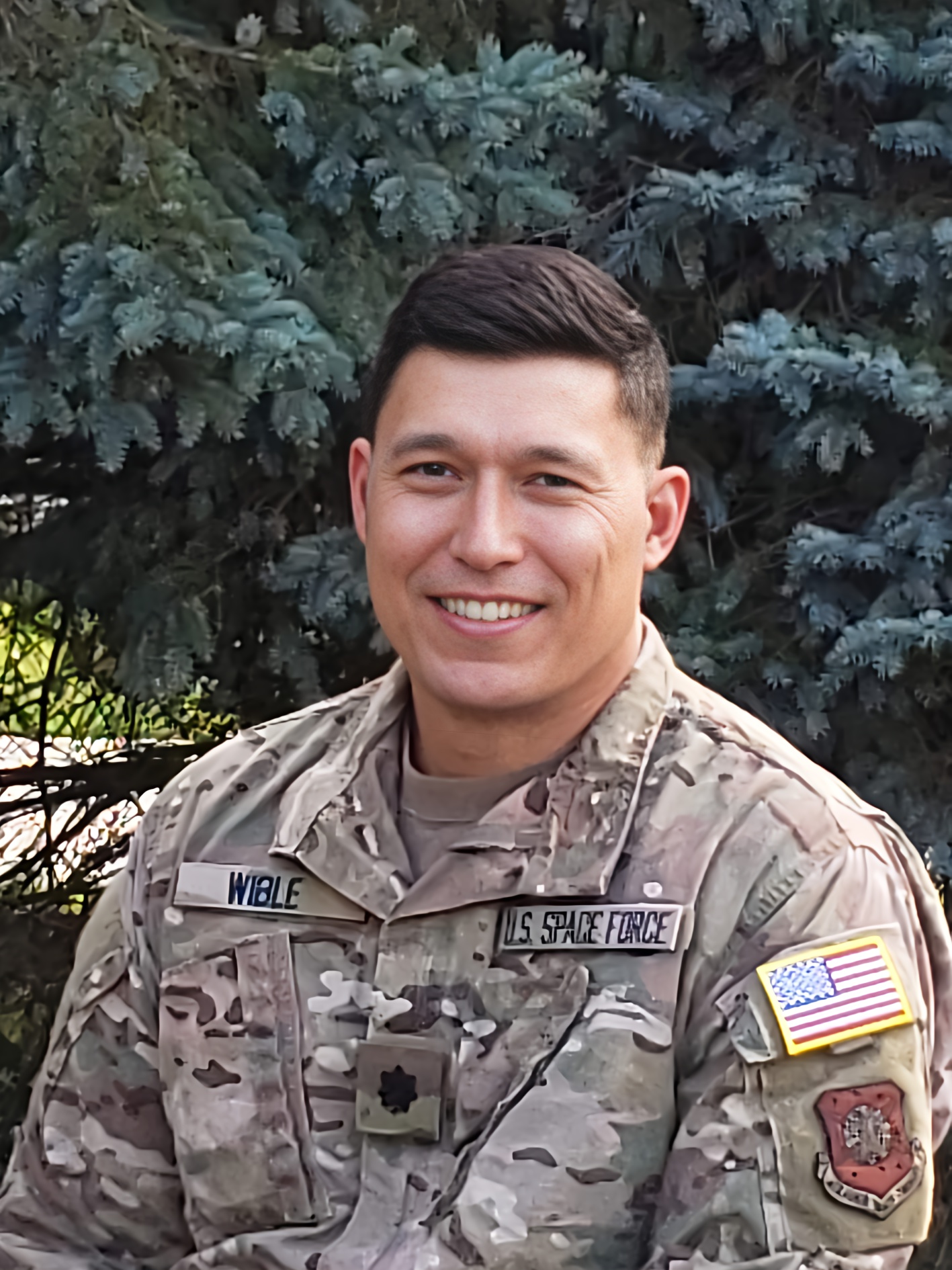}}]{Robert Wible}
Lt. Col. Robert Wible is an active duty officer in the USSF. He obtained a B.S. in Mechanical Engineering and a Ph.D. in Imaging Science from the Rochester Institute of Technology (RIT) in 2005 and 2025, respectively. He also completed an M.S. in Systems Engineering at George Washington University in 2014. He is an engineer with experience in flight test and evaluation, intelligence, and space systems. He has served as chief engineer and program manager for the National Geospatial-Intelligence Agency’s (NGA) advanced analytics portfolio, focusing on activity-based intelligence (ABI), multi-INT data fusion, and artificial intelligence (AI). At Space Systems Command (SSC), he led and negotiated a \$1B joint program with Space Norway to deliver satellite communications to the Polar Regions. During his time at RIT’s Center for Imaging Science, he investigated lidar’s ability to detect objects beneath forest canopies. Under Dr. Jan van Aardt, he designed and built a large-scale, realistic forest scene called “Harvard Forest,” capable of simulating various types of remote sensing data. Using this scene, he developed a novel method to train AI/ML models with simulated data to classify real full-waveform lidar. He is currently serving at the National Reconnaissance Office (NRO) as the Electro-Optical Division Chief for the Geospatial Intelligence (GEOINT) Program Office Directorate.
\end{IEEEbiography}
\begin{IEEEbiography}[{\includegraphics[width=1in,height=1.25in, clip,keepaspectratio]{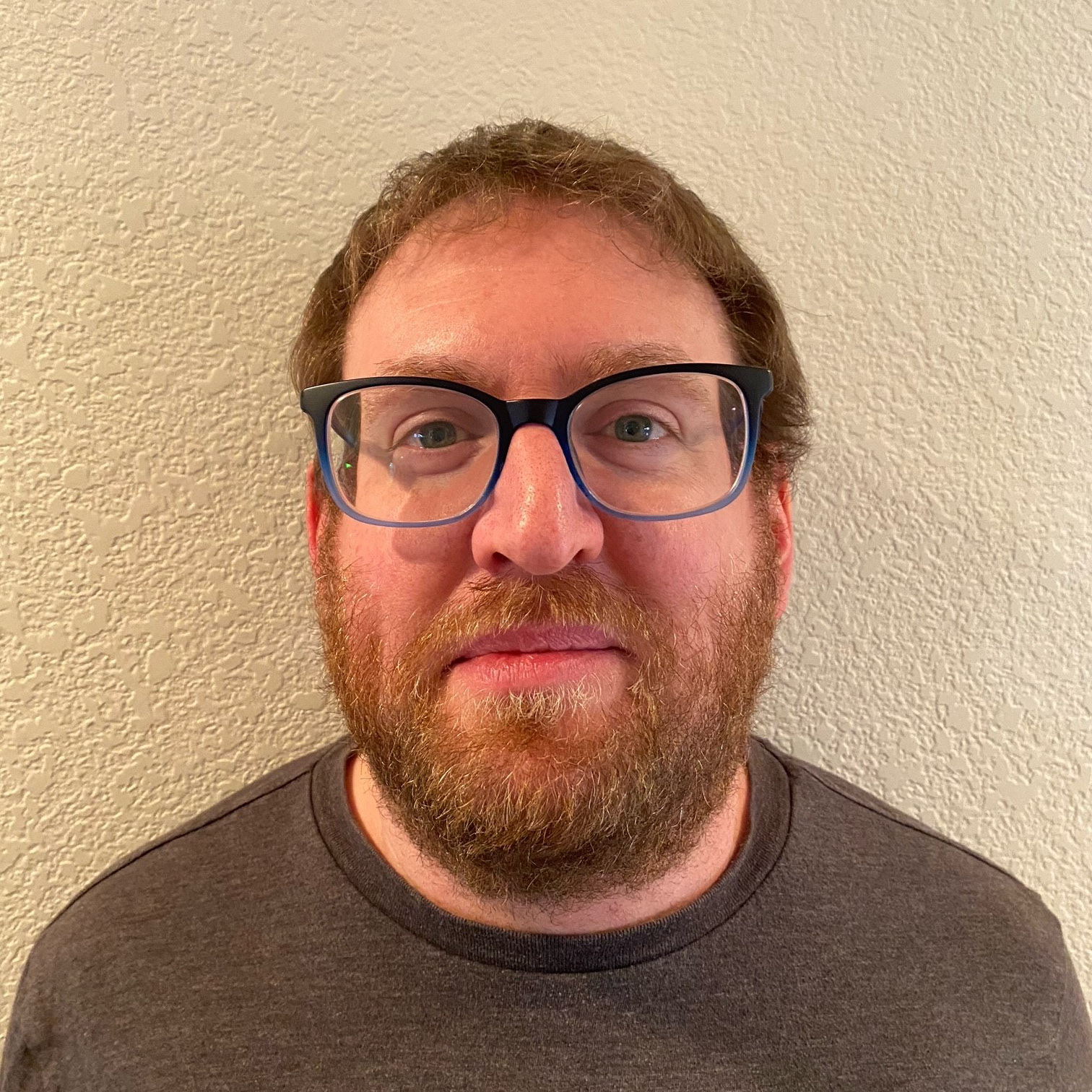}}]{Keith Krause} is an Environmental Scientist at Battelle with a focus on remote sensing of forest ecosystems. He obtained a B.S. in Imaging and Photographic Technology from Rochester Institute of Technology in 1999, a M.S. in Optical Sciences from the University of Arizona in 2001, and a Ph.D. in Aerospace Engineering Sciences from the University of Colorado in 2015. For the first part of his career, Keith performed the radiometric calibration of several imaging satellites while working at DigitalGlobe. Next, he switched to airborne remote sensing as a member of the Science Team at the National Ecological Observatory Network (NEON). More recently, Keith has been the Principal Investigator on several NASA research grants focusing on using lidar to measure vegetation structure as well as understanding radiative transfer in vegetation canopies.  
\end{IEEEbiography}
\begin{IEEEbiography}[{\includegraphics[width=1in,height=1.25in, clip,keepaspectratio]{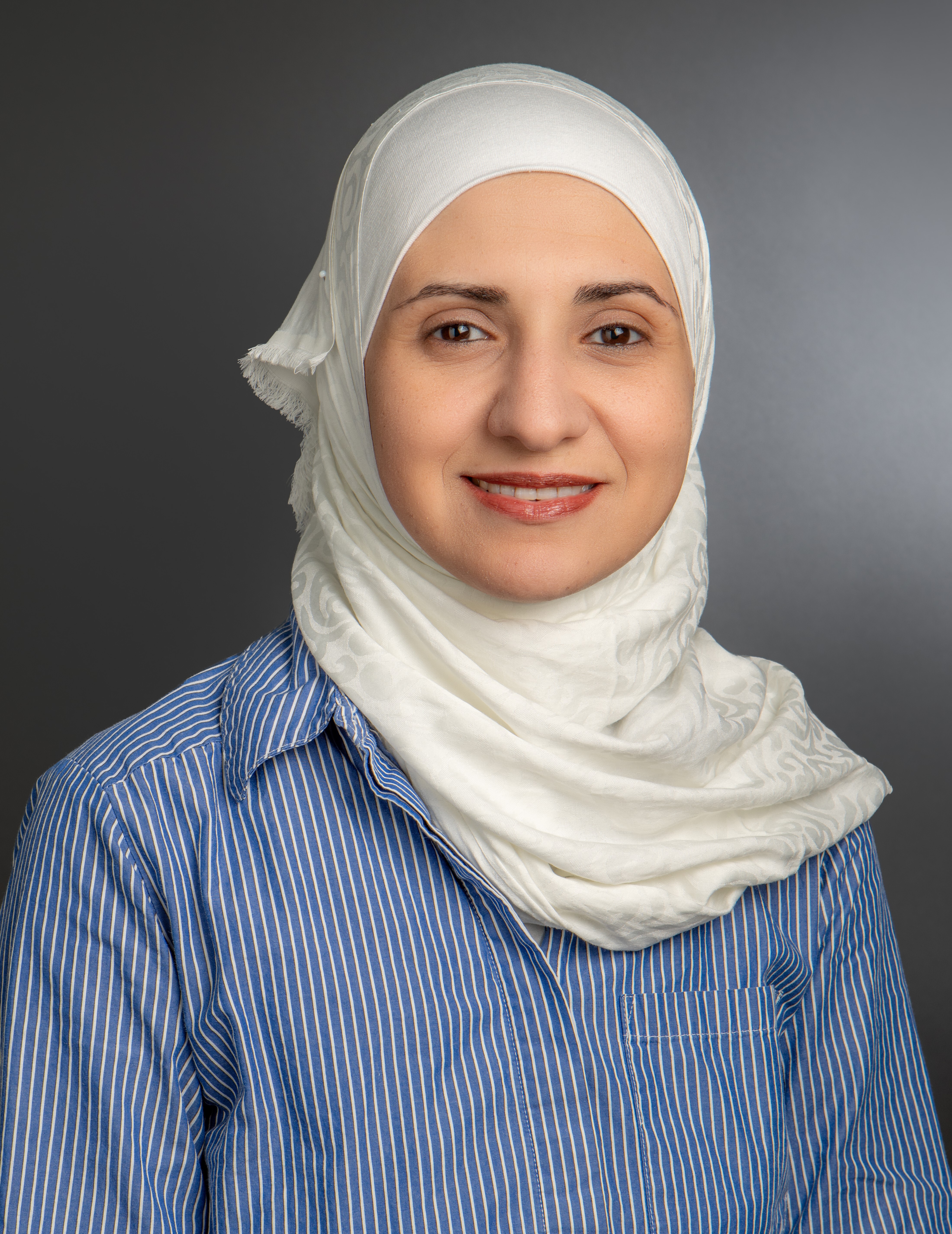}}]{Dimah Dera} is an Endowed Assistant Professor at the Chester F. Carlson Center for Imaging Science at the Rochester Institute of Technology. She received her Ph.D., M.S. in Electrical and Computer Engineering, and M.A. in Mathematics from Rowan University. Dimah received the National Science Foundation (NSF) Computer and Information Science and Engineering Research Initiation Initiative (CRII) Award No. 2401828 in 2023 and the NSF Research Experiences for Undergraduates (REU) supplement award in 2024 for her current research, which focuses on robust and trustworthy machine learning. She won several research Awards at IEEE conferences and the Engineering community, such as the Best Paper Award at the 2019 IEEE International Workshop on Machine Learning for Signal Processing and the IEEE Philadelphia Sections Benjamin Franklin Key Award (2021). Dimah has served as a member of the IEEE Signal Processing and Computational Intelligence Societies as well as a member of the ACM SIGHPC Association for Computing Machinery. She is the NVIDIA Deep Learning Institute (DLI) University Ambassador. Dimah specializes in robust and trustworthy modern machine learning (ML) solutions for real-world applications, including healthcare, remote sensing, and surveillance systems. 
\end{IEEEbiography}
\begin{IEEEbiography}[{\includegraphics[width=1in,height=1.25in, clip,keepaspectratio]{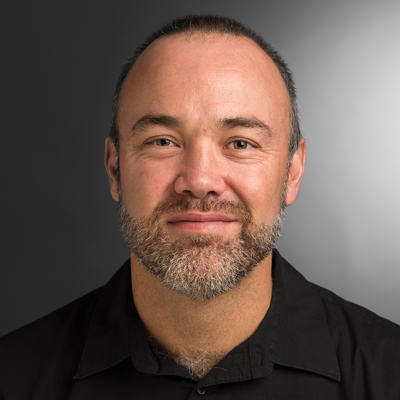}}]{Jan van Aardt} is a professor in the Chester F. Carlson Center for Imaging Science at the Rochester Institute of Technology, New York. He obtained a B.S. Forestry degree (“how to grow and cut down trees”) from the University of Stellenbosch, South Africa in 1996. He completed M.S. and Ph.D. Forestry degrees focused on remote sensing (imaging spectroscopy and light detection and ranging), at the
Virginia Polytechnic Institute and State University, Blacksburg, Virginia in 2000 and 2004, respectively. This was followed by post-doctoral work at the Katholieke
Universiteit Leuven, Belgium, and a stint as a research group leader at the Council for Scientific and Industrial Research, South Africa. Imaging spectroscopy and structural (lidar) sensing of natural resources form the core of his efforts, which vary between vegetation structural and system state (physiology) assessment. Stated differently, the interaction of photons with leaves is what really excites him. He has received funding from NSF, NASA, Google, and USDA, among others, and has published $>$80 peer-reviewed papers and $>$80 conference contributions.
\end{IEEEbiography}




\end{document}